\documentclass{article} 
\usepackage{iclr2027_conference,times}

\usepackage{amsmath,amsfonts,bm}

\def\eqref#1{equation~\ref{#1}}

\def\1{\bm{1}}

\DeclareMathAlphabet{\mathsfit}{\encodingdefault}{\sfdefault}{m}{sl}
\SetMathAlphabet{\mathsfit}{bold}{\encodingdefault}{\sfdefault}{bx}{n}

\usepackage{hyperref}
\usepackage{url}

\iclrfinalcopy

\title{Historical Gradient Reconstruction \\ for Adaptive Federated Continual Learning}

\author{Longtao Xu, Jian Li \\
Stony Brook University\\
\texttt{\{longtao.xu, jian.li.3\}@stonybrook.edu} 
}

\newcommand{\Adaptive}{\textsc{FedGRIT}\xspace} 

\usepackage{amsmath, amsthm, amssymb}
\usepackage{amsfonts} 
\usepackage{booktabs}
\usepackage{multirow}

\newtheorem{assumption}{Assumption}

\newtheorem{theorem}{Theorem}

\newtheorem{corollary}{Corollary}

\newtheorem{lemma}{Lemma}

\newtheorem{remark}{Remark}

\usepackage{color}
\usepackage[table]{xcolor}
\usepackage{makecell}
\usepackage{xcolor}
\usepackage{xspace}
\usepackage{graphicx}
\usepackage{algorithm}
\usepackage{algorithmic}
\usepackage{tcolorbox}
\usepackage{wrapfig}

\definecolor{mygreenbg}{RGB}{238,250,235}
\definecolor{mygreenborder}{RGB}{70,85,70}

\tcbuselibrary{skins,breakable}

\usepackage{caption}

\DeclareCaptionFormat{algline}{%
  \hrule height 0.6pt\relax\par\vspace{2pt}%
  #1#2#3\par\vspace{2pt}%
  \hrule height 0.6pt\relax%
}

\begin{document}

\maketitle
\lhead{Preprint}

\begin{abstract}
Federated continual learning (FCL) must balance the acquisition of new knowledge with the retention of earlier knowledge distributed across heterogeneous clients. Gradient-based task weighting offers a way to adapt this balance, but choosing the weights requires historical gradients from data distributed across clients. Private replay buffers provide only a local view of past data, while shared historical gradients can become inaccurate as clients update their models. We propose \Adaptive (Federated Gradient Reconstruction for Incremental Training), a replay-based framework that reconstructs historical gradients to guide adaptive local updates. By aggregating task-specific gradients and compact curvature information, 
\Adaptive enables clients to \textit{reconstruct historical gradients at their own changing models and adapt task weights without cross-client communication at every step.} 
Periodic refreshes and two-stage curvature compression support repeated sharing.  
We derive conditional historical-loss retention and weighted stationarity bounds under partial participation, separating the effects of stochasticity, local training, and  client heterogeneity.
Across four class- and domain-incremental benchmarks, \Adaptive achieves the highest mean final accuracy among the evaluated methods with ablations examining shared reconstruction and adaptive weighting. 
\end{abstract}

\section{Introduction}

Continual learning (CL) studies how models learn sequential tasks with limited access to earlier data. Its central challenge is to balance \textit{stability}, the retention of previously acquired knowledge, and \textit{plasticity}, the acquisition of new knowledge~\citep{LopezPazRanzato2017GEM}. When such learning takes place across clients that keep their data local, federated continual learning (FCL) enables collaboration over evolving private data~\citep{yoon2021federated}. In this paper, we study the shared-model setting, where local updates must support new-task learning while preserving historical knowledge distributed across heterogeneous clients.

{
Existing FCL methods draw on distillation or regularization~\citep{usmanova2022federated,li2025fedssi}, task-specific parameter sharing~\citep{yoon2021federated}, and replay~\citep{keshri2025convergence} to retain earlier knowledge. A complementary challenge is to determine how strongly current and historical tasks should influence each local update. Their gradients may agree or conflict, and these relationships can change across clients and training stages.

Gradient-based adaptive weighting responds to these changes using the norms and alignment of task gradients. Pareto-based CL~\citep{wu2024mitigating} and federated multi-objective optimization~\citep{yang2023federated,askin2025federated} provide foundations for this approach, while gradient correction has also been studied directly in FCL~\citep{zhang2025fedagc}. The effectiveness of such decisions, however, depends on the historical-gradient estimates available to each client.

A private replay buffer captures only a local view of the federation's history: an update compatible with one client's retained samples may still interfere with knowledge held elsewhere. Sharing replay gradients broadens this view. For example, C-FLAG~\citep{keshri2025convergence} shares a pooled historical gradient at the round's shared model and adapts the balance between memory and current-data updates. However, a pooled gradient at the round boundary does not provide separate historical-task estimates at subsequent local models. Adaptive per-task decisions require both distinctions: which historical task is affected and how its gradient changes along the receiving client's local trajectory. Re-evaluating these gradients across clients after every local step would require repeated communication. This raises our central question:}
\vspace{-0.1in}
\begin{tcolorbox}[colback=white!5!white,colframe=white!75!white]
\textit{How can distributed historical information support adaptive task balancing throughout local training without requiring cross-client communication at every update?}
\end{tcolorbox}
\vspace{-0.1in}

{
We answer this question with \Adaptive (Federated Gradient Reconstruction for Incremental Training), a replay-based framework that couples task-specific reconstruction of distributed historical gradients with adaptive local weighting.} Building on past-gradient reconstruction in CL~\citep{daxberger2023improving}, participating clients contribute task-specific replay gradients and compact curvature information at a common model. Aggregation incorporates history from other clients, while curvature corrections adjust the estimates as local models move. 
Each client combines these reconstructed gradients with its current-task gradient to adapt task weights at every local step, without further cross-client queries. Periodic refreshes recenter the summaries, while client compression and server recompression keep each curvature summary compact.

{
Building on historical-gradient reconstruction \citep{daxberger2023improving} and gradient-based task balancing \citep{wu2024mitigating,yang2023federated}, \Adaptive addresses their coupling under distributed historical-data access. Clients exchange task-specific summaries at a common model, then reconstruct historical gradients and adapt task weights along their own local trajectories. This separates historical-information exchange from per-step adaptation without restricting each client's decisions to its own replay buffer.}

Our main contributions are as follows:

$\bullet$ \textbf{Shared reconstruction for adaptive local updates.} We develop \Adaptive, which combines task-specific historical gradients across clients and adjusts them to each evolving local model, enabling adaptive task weighting without cross-client queries at every local step. 

$\bullet$ \textbf{Conditional historical-loss retention.} We bound the increase in cumulative historical-task loss under an alignment condition on the adaptive local directions, explicitly separating stochastic noise and client sampling.

$\bullet$ \textbf{Weighted stationarity under local adaptation.} We analyze the federation-wide task-gradient combination induced by client-specific local weights, distinguishing optimization and local-update effects from sampling-related and persistent client heterogeneity.

$\bullet$  \textbf{Evaluation across incremental settings.} \Adaptive achieves the highest mean final accuracy among the evaluated methods on four class- and domain-incremental benchmarks. Component and sensitivity studies examine shared reconstruction, weighting, and curvature sharing.

\section{Problem Setting}\label{sec:setting}

We consider a server coordinating $M$ clients to train a shared model. Clients follow a shared task sequence with synchronized task transitions~\citep{keshri2025convergence,xu2026serverproximal}, while their data distributions may differ within each task. 
{ Synchronization provides common task identities and a common current-historical task split, but it does not require identical client data, gradients, or curvature. We study within-task data heterogeneity, rather than client-specific task orders or unsynchronized task transitions.}
At stage $K$, client $m$ observes its current-task dataset $\mathcal T_{K,m}$ and maintains a private replay buffer of fixed total capacity containing samples from earlier tasks. Earlier-task data are accessible only through these buffers, and raw samples are not exchanged between clients or with the server. Replay samples retain their task identities for training.

For model parameters $\theta\in\mathbb R^d$ and task $i$, let $f_{i,m}(\theta)=\mathbb E_{(x,y)\sim\mathcal T_{i,m}}[f_{i,m}(\theta;x,y)]$ denote the loss of client $m$, where $(x,y)$ is an input-label pair. The federation-wide task loss is $f_i(\theta)=M^{-1}\sum_{m=1}^{M}f_{i,m}(\theta)$. The cumulative learning objective is
\begin{equation}
  \min_{\theta\in\mathbb R^d} 
  f_{1:K}(\theta):=\sum_{i=1}^{K}f_i(\theta)=\frac{1}{M}\sum_{i=1}^{K}\sum_{m=1}^{M}f_{i,m}(\theta).
  \label{eq:revised-objective}
\end{equation}
This objective accounts for both learning the current task and retaining knowledge from earlier tasks. During new-task training, however, the full earlier-task datasets are unavailable, and each client's replay buffer contains only a limited sample of its own history. \textit{An update that preserves this local history may therefore still degrade performance on historical data held by other clients.}

At communication round $t$, the server samples a set $\mathcal S_K^t$ of $N$ clients uniformly without replacement. We use $T$ for the number of rounds per task and $E$ for the number of local gradient steps performed by each participating client per round. The shared model at the start of round $t$ is $\theta_K^t$, and $\theta_{K,m}^{t,e}$ denotes client $m$'s model after $e$ local steps, with $\theta_{K,m}^{t,0}=\theta_K^t$. For $K>1$, training on task $K$ starts from $\theta_K^0=\theta_{K-1}^{T}$. The local and server learning rates are $\eta_L$ and $\eta_G$, respectively.

\begin{figure}[t]
    \centering
    \includegraphics[width=0.9\linewidth]{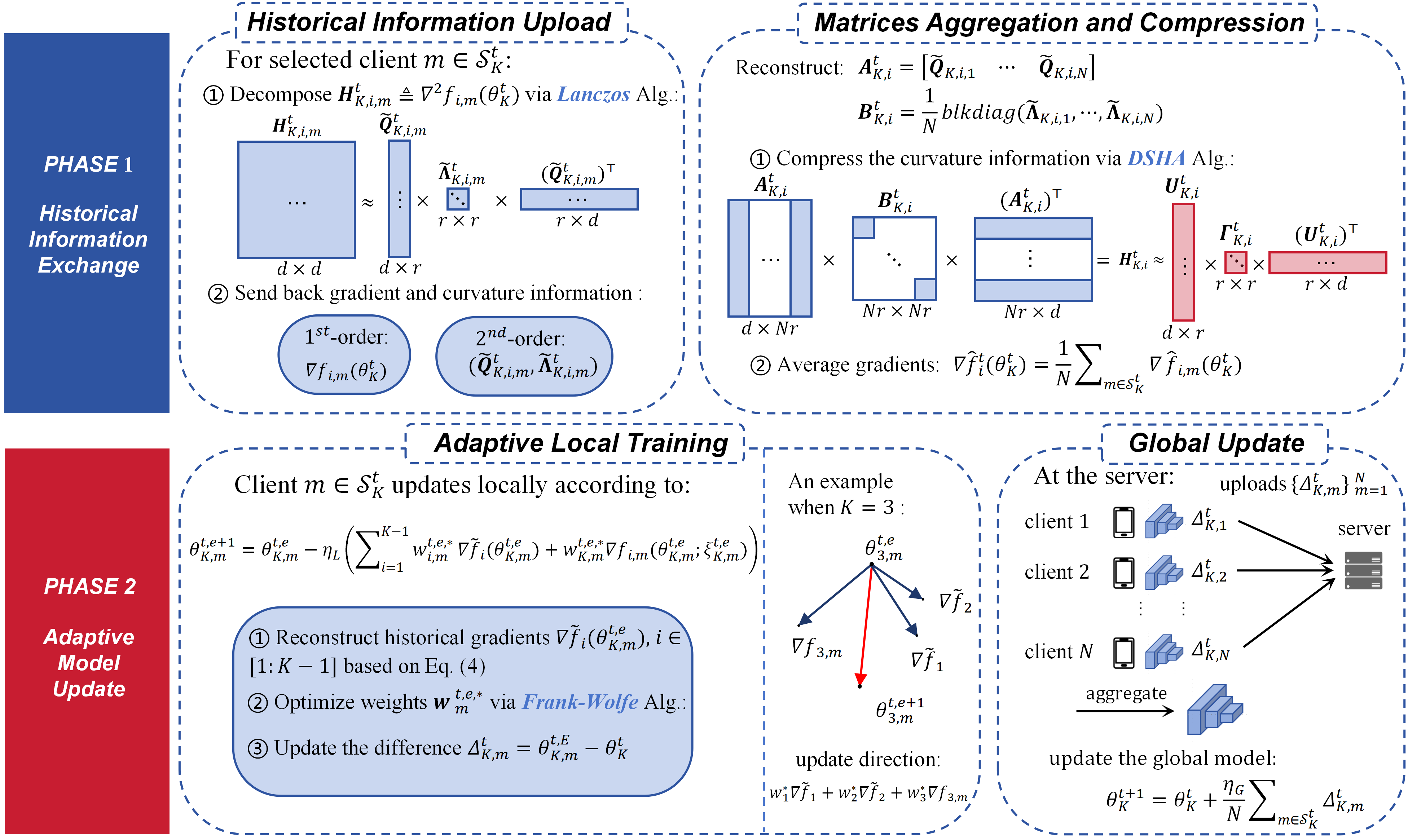}
\caption{\textbf{\Adaptive in one communication round.} Clients share task-specific replay gradients and low-rank curvature factors, which the server aggregates, recompresses, and broadcasts. During \(E\) local steps, each client reconstructs historical gradients at its current model and adapts task weights without further communication. The server then aggregates the model updates.
}
\vspace{-0.1in}
    \label{fig:workflow}
\end{figure}

\section{\Adaptive: Adaptive Updates from Reconstructed History}
\label{sec:algorithm}

Gradient-based task balancing coordinates current- and previous-task gradients to support new learning while limiting interference with earlier knowledge~\citep{wu2024mitigating}. In FCL, each client learns from its current-task data, while the historical knowledge to be retained is distributed across clients. Its local update should therefore account for how learning from its current data affects earlier tasks across the federation. For client $m$ training on task $K$, we consider the following idealized direction at local step $e$ of communication round $t$:
\begin{equation}\label{eq:truly local update}
    \boldsymbol{\mathrm{d}}_{K,m}^{t,e} = \sum_{i=1}^{K-1}w_{i,m}^{t,e} \nabla {f}_i(\theta_{K,m}^{t,e}) + w_{K,m}^{t,e}\nabla f_{K,m}(\theta_{K,m}^{t,e};\xi_{K,m}^{t,e}).
\end{equation}
Here, $\nabla f_i(\theta_{K,m}^{t,e})$ is the federation-wide gradient of historical task $i$, evaluated at client $m$'s current local model. The current-task gradient $\nabla f_{K,m}(\theta_{K,m}^{t,e};\xi_{K,m}^{t,e})$ is computed from a  minibatch $\xi_{K,m}^{t,e}$ drawn from client $m$'s current-task data. The weights $\boldsymbol{w}_m^{t,e}
=(w_{1,m}^{t,e},\ldots,w_{K,m}^{t,e})^\top$ determine each task's contribution, with $w_{i,m}^{t,e}\geq 0$ and $\sum_{i=1}^{K}w_{i,m}^{t,e}=1$.

Constructing this direction requires historical gradients at the changing local model and weights that balance their contributions. \textit{These requirements are coupled: adapting the weights using gradient agreement requires historical-gradient estimates that remain informative throughout local training.} A private replay buffer provides only a limited view of the federation's history. Sharing replay gradients can broaden this view~\citep{keshri2025convergence}, but gradients evaluated at the shared model can become inaccurate as a client's local model changes. Recomputing the shared replay gradients at every local iterate would require repeated communication with other clients.
\Adaptive addresses this difficulty through shared historical-gradient reconstruction.

\subsection{Reconstructing Historical Gradients During Local Training}
\label{sec:gradient reconstruction}

{
Adaptive weighting compares historical and current gradients at a client's local model. However, a shared historical gradient is evaluated at $\theta_K^t$, while the current-task gradient changes during local training. Reusing the shared gradient can therefore distort the task relationships used for weighting. \Adaptive uses curvature to approximate how the historical gradient changes, allowing different clients to adjust the shared information along their own local trajectories. Specifically, we use the first-order Taylor approximation of the gradient}
\begin{equation}\label{eq:first order expansion}
    \nabla f_i(\theta_{K,m}^{t,e})\approx \nabla f_i(\theta_K^t)+\boldsymbol{H}_{K,i}^t(\theta_{K,m}^{t,e}-\theta_K^t),
\end{equation}
where $\boldsymbol{H}_{K,i}^t:=\nabla^2 f_i(\theta_K^t)$ is the federation-wide task-$i$ Hessian at the shared model. The curvature term estimates how the historical gradient changes as client $m$ moves away from $\theta_K^t$. The exact gradient and Hessian in this approximation are unavailable because the full historical datasets are no longer accessible.

We estimate these quantities using participating clients' private replay buffers. During training on task $K$, let $\widehat f_{i,m}$ denote the empirical loss on client $m$'s retained samples from task $i$. For the participating-client set $\mathcal S_K^t$, define the
shared replay loss $\widehat f_i^t(\theta):=\frac{1}{N}\sum_{m\in\mathcal S_K^t}\widehat f_{i,m}(\theta).$
Its gradient at $\theta_K^t$ is obtained by averaging the replay gradients uploaded by participating clients. It estimates the federation-wide gradient using their retained data. Let $\tilde{\boldsymbol{H}}_{K,i}^t$ denote the compact curvature approximation constructed from the same clients' replay losses, as detailed in Section~\ref{sec:workflow}.
Each client then reconstructs the historical gradient as
\begin{equation}\label{eq:reconstructed direction}
    \nabla\tilde f_i(\theta_{K,m}^{t,e}):={ \nabla\widehat f_i^t(\theta_K^t)}+\tilde{\boldsymbol{H}}_{K,i}^t(\theta_{K,m}^{t,e}-\theta_K^t).
\end{equation}
The shared gradient incorporates history from participating clients, while the curvature term adjusts it to the receiving client’s model. Setting the curvature term to zero recovers reuse of the shared gradient. 
{
The reconstructed gradients are then used to select task weights at each local step, allowing adaptive weighting to account for the estimated changes in historical gradients during local training.}
Accuracy depends on replay coverage, client sampling, curvature approximation, and deviation from the shared model.

\subsection{Adaptive Task Weighting and Local Updates}
\label{sec:adaptive weighting}

Reconstruction supplies historical-gradient estimates at the client's current local model. We use these estimates together with the current-task gradient to adjust task weights according to their norms and mutual alignment. Following gradient-based multi-objective optimization~\citep{yang2023federated}, we select a minimum-norm convex combination:
\begin{equation}\label{eq:adaptive task weights}
\boldsymbol{w}_{m}^{t,e,*}\in\underset{\substack{\boldsymbol{w}\geq\boldsymbol{0}\\\sum_{i=1}^{K}w_i=1}}{\arg\min} \left\|\sum_{i=1}^{K-1}w_i\nabla\tilde f_i(\theta_{K,m}^{t,e})+w_K\nabla f_{K,m}(\theta_{K,m}^{t,e};\xi_{K,m}^{t,e})\right\|^2.
\end{equation}
Because the current-task weight is also optimized, the rule adjusts both the relative emphasis among historical tasks and the balance between historical and current learning.
{
At an exact solution, a nonzero minimum-norm direction has positive inner product with every supplied gradient at the current local model. This local geometric property does not automatically imply descent for the true task losses or survive server aggregation. Client-specific, per-step weighting allows the task balance to respond to heterogeneous current-task data and evolving local gradient relationships.
}

For fixed gradient estimates, Eq.~(\ref{eq:adaptive task weights}) is a convex quadratic problem in $K$ weights, approximately solved by Frank–Wolfe (Appendix~\ref{sec:frank wlfe algorithm}~\citep{frank1956algorithm}). We write the updates using its exact minimizer $\boldsymbol{w}_{m}^{t,e,*}$ for the analysis: 
\begin{equation}\label{eq:adaptive local direction}
\tilde{\boldsymbol{\mathrm{d}}}_{K,m}^{t,e,*}=\sum_{i=1}^{K-1}w_{i,m}^{t,e,*}\nabla\tilde f_i(\theta_{K,m}^{t,e})+w_{K,m}^{t,e,*}\nabla f_{K,m}(\theta_{K,m}^{t,e};\xi_{K,m}^{t,e}),
\end{equation}
\begin{equation}\label{eq:adaptive local update}
\theta_{K,m}^{t,e+1}=\theta_{K,m}^{t,e}-\eta_L\tilde{\boldsymbol{\mathrm{d}}}_{K,m}^{t,e,*}.
\end{equation}

The shared summaries remain fixed within a round, while reconstruction and weighting are repeated at every local step. The curvature correction adjusts the historical-gradient estimates as the local model moves, and the weights respond to the resulting relationships among current and historical gradients. Both computations are performed locally without further cross-client communication. For the first task, the historical sum is empty and $w_{1,m}^{t,e,*}=1$.

\subsection{Sharing Compact Curvature Information}
\label{sec:workflow}

Historical gradients can be averaged directly, but a full client Hessian contains $d^2$ entries. \Adaptive therefore compresses curvature before upload and recompresses the aggregate before broadcast.

\textbf{Client-side compression.}
For each historical task $i<K$, participating client $m$ approximates the Hessian of its replay loss at the shared model:
\begin{equation}\label{eq:client curvature compression}
\nabla^2\widehat f_{i,m}(\theta_K^t)\approx\tilde{\boldsymbol{Q}}_{K,i,m}^t\tilde{\boldsymbol{\Lambda}}_{K,i,m}^t(\tilde{\boldsymbol{Q}}_{K,i,m}^t)^\top,
\end{equation}
Here, $\widetilde{\boldsymbol{Q}}_{K,i,m}^t \in \mathbb{R}^{d\times r}$ has orthonormal columns, and $\widetilde{\boldsymbol{\Lambda}}_{K,i,m}^t \in \mathbb{R}^{r\times r}$ is diagonal with the corresponding signed eigenvalue estimates. We use Lanczos with Hessian-vector products to approximate the $r$ eigenpairs with largest eigenvalue magnitudes, without forming the full Hessian~\citep{lanczos1950iteration,chen2021analysis}. The procedure is given in Algorithm~\ref{alg:Hessian-Lanczos Decomposition} in Appendix~\ref{sec:lanczos algorithm}. Each client uploads these factors together with its replay gradient $\nabla\widehat{f}_{i,m}(\theta_K^t)$. { At each refresh, Lanczos uses up to $k$ HVPs per historical task and participating client, while the retained rank $r\leq k$ controls factor storage and communication. Local reconstruction then applies the cached factors in $\mathcal{O}(dr)$ operations per historical task per step, without additional HVPs.}

\textbf{Server-side aggregation and recompression.}
Averaging rank-$r$ client approximations can produce rank up to $\min(Nr,d)$, so broadcasting all collected factors would make each recipient's curvature payload grow with the number of participants. Distributed Spectral Hessian Aggregation (DSHA) recompresses this aggregate by retaining at most $r$ eigenpairs with the largest eigenvalue magnitudes, preserving their signs:
\begin{equation}\label{eq:server curvature recompression}
\tilde{\boldsymbol{H}}_{K,i}^t:=\boldsymbol{U}_{K,i}^t\boldsymbol{\Gamma}_{K,i}^t(\boldsymbol{U}_{K,i}^t)^\top\approx\frac{1}{N}\sum_{m\in\mathcal S_K^t}\tilde{\boldsymbol{Q}}_{K,i,m}^t\tilde{\boldsymbol{\Lambda}}_{K,i,m}^t(\tilde{\boldsymbol{Q}}_{K,i,m}^t)^\top.
\end{equation}
The columns of $\boldsymbol{U}_{K,i}^t\in\mathbb{R}^{d\times r}$ are orthonormal, and $\boldsymbol{\Gamma}_{K,i}^t\in\mathbb{R}^{r\times r}$ contains the retained signed eigenvalues. { The QR-based construction and DSHA pseudocode are provided in Appendix~\ref{sec:Distributed Spectral Hessian Aggregation Algorithm}.}
{
DSHA uses standard QR factorization and truncated eigendecomposition to recompress the aggregated curvature into a compact shared representation. It operates on the span of the uploaded client subspaces and does not require a common eigenspace. The QR step changes coordinates without altering the represented aggregate, whereas client-side compression and server-side spectral truncation introduce approximation error. Components absent from the uploaded span cannot be recovered, even if they are important in the uncompressed aggregate. The retained rank therefore controls a tradeoff between curvature fidelity and communication cost.}

The server broadcasts the averaged replay gradient $\nabla\widehat{f}_i^t(\theta_K^t)$ and the factors $(\boldsymbol{U}_{K,i}^t,\boldsymbol{\Gamma}_{K,i}^t)$ for each historical task. Clients apply these factors directly in Eq.~(\ref{eq:reconstructed direction}), without forming a $d\times d$ matrix, at a cost of $\mathcal{O}(dr)$ per historical task per local step. For $r\geq 1$, each client receives $\mathcal{O}((K-1)dr)$ scalar entries per historical-summary refresh. Recompression makes this per-client payload independent of $N$, although it still grows with the number of historical tasks.

{
Only the $N$ selected clients contribute summaries at each refresh. For fixed model size, task count, rank, and local replay workloads, increasing the total population $M$ while keeping $N$ fixed does not increase the curvature-exchange workload per refresh. Increasing $N$, however, increases aggregate traffic and server processing, even though each recipient's summary size remains unchanged.}

\textbf{Complete round protocol.}
Algorithm~\ref{alg:fedgrit complete} summarizes \Adaptive with historical summaries refreshed at the beginning of every round. After exchanging summaries at $\theta_K^t$, participating clients perform $E$ adaptive local steps as described in Section~\ref{sec:adaptive weighting}, without further cross-client queries. Each client uploads
$\Delta_{K,m}^t=\theta_{K,m}^{t,E}-\theta_K^t$,
and the server updates
\begin{equation}\label{eq:fedgrit server update}
\theta_K^{t+1}=\theta_K^t+\eta_G\Delta_K^t, \qquad \Delta_K^t:=\frac{1}{N}\sum_{m\in\mathcal S_K^t}\Delta_{K,m}^t.
\end{equation}
{ At each task boundary, each client reallocates its fixed total replay budget approximately equally among the observed tasks. For earlier tasks, it uniformly subsamples without replacement from the retained samples. For the newly completed task, it uniformly samples without replacement from its local data. The total replay budget remains fixed as new tasks arrive.}
{
Raw samples remain local throughout the procedure, but data locality does not constitute a formal privacy guarantee. The server observes each participating client's task-specific replay gradients and curvature factors before aggregation, while clients receive only aggregated summaries. Repeated releases from retained samples may enable inference about local data. Neither low-rank compression nor ordinary aggregation provides protection against such attacks. Additional privacy mechanisms would need to cover both gradient and curvature information.}

\begin{algorithm}[t]
\caption{\Adaptive for task $K$ with per-round summary refresh}
\label{alg:fedgrit complete}
\begin{algorithmic}[1]
\REQUIRE Initial model $\theta_K^0$, clients' current-task datasets and private replay buffers, $T,E,N,r,\eta_L,\eta_G$
\ENSURE Shared model $\theta_K^T$ and updated private replay buffers
\FOR{$t=0,\ldots,T-1$}
  \STATE Server samples $N$ clients as $\mathcal S_K^t$ and sends them $\theta_K^t$.
  \IF{$K>1$}
    \STATE Each participating client computes and uploads replay gradients and Lanczos curvature factors at $\theta_K^t$ for tasks $i=1,\ldots,K-1$.
    \STATE Server averages the replay gradients and recompresses each task's curvature using DSHA.
    \STATE Server broadcasts the shared gradients and curvature factors.
  \ENDIF

  \FOR{each client $m\in\mathcal S_K^t$ in parallel}
    \STATE Initialize $\theta_{K,m}^{t,0}=\theta_K^t$.
    \FOR{$e=0,\ldots,E-1$}
      \STATE Reconstruct historical-task gradients using Eq.~(\ref{eq:reconstructed direction}) when $K>1$.
      \STATE Draw $\xi_{K,m}^{t,e}$ and compute the current-task gradient.
      \STATE For $K>1$, approximately solve Eq.~(\ref{eq:adaptive task weights}) using Frank-Wolfe;
      for $K=1$, set $w_{1,m}^{t,e,*}=1$.
      \STATE Use the computed weights to form the direction and update the local model according to Eqs.~(\ref{eq:adaptive local direction})-(\ref{eq:adaptive local update}).
    \ENDFOR

    \STATE Upload
    $\Delta_{K,m}^t=\theta_{K,m}^{t,E}-\theta_K^t$.
  \ENDFOR

  \STATE Server updates $\theta_K^{t+1}$ using Eq.~(\ref{eq:fedgrit server update}).
\ENDFOR

\STATE Clients update their fixed-size private replay buffers.
\end{algorithmic}
\end{algorithm}

\section{Theoretical Analysis}\label{sec:theoretical analysis}

We analyze how new task training affects cumulative historical loss and how the adaptive updates approach weighted stationarity across the tasks seen so far. 

\subsection{Assumptions}
\begin{assumption}[Bounded Gradients]\label{assumption1:Bounded Gradient}
    For every task  $i\in[K]$ and client $m\in[M]$, the  gradient is bounded by a constant $D\geq 0$, i.e., $\mathbb{E}[\Vert \nabla f_{i,m}(\theta)\Vert]\leq D, \forall \theta\in\mathbb{R}^{d}$.
\end{assumption}

\begin{assumption}[L-smoothness]\label{assumption2:L-smooth}
    For every task  $i\in[K]$, client $m\in[M]$, and all $\theta,\theta'\in\mathbb{R}^d$, there exists a constant $L>0$ such that $\Vert\nabla f_{i,m}(\theta) - \nabla f_{i,m}(\theta')\Vert\leq L\Vert\theta - \theta'\Vert$.
\end{assumption}

\begin{assumption}[Unbiased Stochastic Estimation]\label{assumption3:Unbiased Stochastic Estimation}
    For every task  $i\in[K]$, client $m\in[M]$, the stochastic gradient estimation is unbiased, i.e., $\mathbb{E}_t[\nabla f_{i,m}(\theta, \xi)] = \nabla f_{i,m}(\theta), \forall \xi\in\mathcal{T}_{i,m}$.
\end{assumption}

\begin{assumption}[Bounded Local Variance]\label{assumption4:Bounded Local Gradient Variance}
    For every task  $i\in[K]$, client $m\in[M]$, there always exists a constant $\sigma_L>0$, such that the variance of stochastic gradient is bounded by $\mathbb{E}_t\left\Vert \nabla f_{i,m}(\theta;\xi) - \nabla f_{i,m}(\theta)\right\Vert^2\leq \sigma_L^2$.
\end{assumption}

{ \begin{assumption}[$(L, \beta)$-Bounded Global Heterogeneity]\label{assumption5:Bounded Global Gradient Variance}
    For every task  $i\in[K]$, client $m\in[M]$, and all $\theta,\theta'\in\mathbb{R}^d$,  there exists a constant $\beta>0$, such that the global heterogeneity of the local gradient is bounded by $\left\Vert \nabla f_{i}(\theta) - \nabla f_{i,m}(\theta')\right\Vert^2\leq L^2\Vert\theta - \theta'\Vert^2 + \beta\sigma_G^2$.
\end{assumption}}

{ Assumptions~\ref{assumption1:Bounded Gradient}-\ref{assumption4:Bounded Local Gradient Variance} control gradient magnitude, smoothness, and stochastic estimation. Assumption~\ref{assumption5:Bounded Global Gradient Variance} states that the global-local gradient discrepancy is jointly controlled by the model deviation and client heterogeneity. When $\theta = \theta'$, Assumption~\ref{assumption5:Bounded Global Gradient Variance} reduces to the conventional bounded global variance assumption (assumption 3 in~\citet{yang2021achieving}), and it also reduces to the standard Lipschitz continuous assumption as Assumption~\ref{assumption2:L-smooth} when specialized to the i.i.d. case.}

\subsection{Backward Knowledge Transfer Analysis}

{

We compare the cumulative historical loss during task-$K$ training with its value at the task-start model $\theta_K^0=\theta_{K-1}^{T}$. The following result bounds its increase under an alignment condition on the local update directions.

}

\begin{theorem}[BWT under partial participation]\label{theorem:BKT}

Suppose for each $\tau\in[t]$, local gradient step $e\in[E]$, and client $m\in[M]$, the update direction $\tilde{\boldsymbol{\mathrm{d}}}_{K,m}^{\tau,e}$ is aligned with earlier-task gradient at task-K start, i.e., $\langle\nabla f_{1:K-1}(\theta_K^0), \tilde{\boldsymbol{\mathrm{d}}}_{K,m}^{\tau,e}\rangle\geq \epsilon\Vert\nabla f_{1:K-1}(\theta_K^0)\Vert\cdot\Vert\tilde{\boldsymbol{\mathrm{d}}}_{K,m}^{\tau,e}\Vert, \epsilon\in(0,1)$, and local and global learning rates satisfy $\eta_L\leq\min\{\frac{1}{3(K-1)EL}, \frac{2\epsilon\left\Vert\nabla f_{1:K-1}\left(\theta_K^0\right)\right\Vert}{DLEt\sqrt{7tEN}}\}$ and $\eta_G\leq\frac{1}{K-1}$. Then, for every $t\geq 1$, the server-updated model $\theta_K^t$ generated by the algorithm satisfies:
    \begin{equation}\label{eq:BWT}
        \begin{split}
            \mathbb{E}_t&\left[ f_{1:K-1}\left(\theta_K^t\right)\right]\leq f_{1:K-1}\left(\theta_K^0\right)  + \frac{2\epsilon^2\left\Vert\nabla f_{1:K-1 }\left(\theta_K^0\right)\right\Vert^2}{7\left(K-1\right)Et^2NLD^2}\left({\sigma_L^2} + \frac{\left(M-N\right){ \beta}\sigma_G^2}{N\left(M-1\right)}\right).
        \end{split}
    \end{equation}
\end{theorem}

{
\textbf{Implications.} Eq.~(\ref{eq:BWT}) bounds historical-loss deterioration relative to the task-start model, separating local stochasticity from sampling-related client heterogeneity. Increasing $N$ reduces both the common $1/N$ factor and the sampling factor. Full participation eliminates the heterogeneity contribution but not necessarily the stochastic contribution. The guarantee requires alignment with the task-start historical gradient, and its dependence on $t$ and $E$ must be interpreted together with the corresponding step-size restrictions. It controls cumulative historical loss, not improvement on every earlier task.

}

\begin{remark}[Relation to prior backward-transfer guarantees]
    
{\citet{lin2022beyond} show that, under gradient alignment and a suitable step-size restriction, learning a new task does not increase an earlier-task loss (Theorem~2). Our alignment condition follows the same principle, applied to adaptive local update directions and the cumulative historical loss in a federated setting. From a complementary perspective, \citet{shi2023unified} bound the error on a previous domain by the historical model's error plus a prediction-disagreement term (Lemma~3.2). Equation~(\ref{eq:BWT}) also compares historical performance before and after new-task learning, but expresses the correction through stochasticity and client sampling in the federated updates. \citet{xu2026serverproximal} study historical-loss retention under server-proximal aggregation, with an explicit stochastic correction proportional to $\sigma_L^2$ in its BWT bound. Here, the guarantee concerns adaptive local directions and additionally separates the contribution of $\beta\sigma_G^2$ through the factor $(M-N)/(N(M-1))$. This distinction identifies how partial participation enters the historical-loss bound for heterogeneous clients.}
\end{remark}

\subsection{Weighted task-uniform convergence of \Adaptive}

{
The preceding BWT analysis concerns changes in historical loss. We now examine stationarity across the tasks seen so far. Since \Adaptive adjusts task weights separately at each client's local steps, we evaluate the resulting combination of federation-wide task gradients: $\nabla \tilde{f}_{1:K}(\theta_K^t) \triangleq \sum_{i=1}^{K} w_{i}^{t,*} \nabla f_i(\theta_K^t)$, where $w_{i}^{t,*}\triangleq\frac{1}{EN}\sum_{m\in\mathcal{S}_K^t}\sum_{e=0}^{E-1} w_{i,m}^{t,e,*}$ is the average weight assigned to task $i$ across participating clients and local steps in round $t$. These weights are nonnegative and sum to one, so the resulting gradient combination provides a weighted stationarity criterion for tasks $1,\ldots,K$. A zero value certifies Pareto stationarity,  though individual task gradients need not vanish.}

\begin{theorem}[Weighted Task-Uniform Convergence of \Adaptive]\label{theorem:convergence}
    Suppose Assumptions \ref{assumption1:Bounded Gradient}-\ref{assumption5:Bounded Global Gradient Variance} hold, 
    local and global learning rates $\eta_L$ and $\eta_G$ satisfy $\eta_L\leq \frac{1}{3\left(K-1\right)EL}, K\geq3$ , and $\eta_G\eta_L\leq \frac{1}{6EL}$. Then, the outputs of \Adaptive satisfy
    \begin{align}\label{eq:weighted convergence}
        \frac{1}{T}\sum_{t=0}^{T-1}\mathbb{E}_t\left\Vert\nabla \tilde{f}_{1:K}\left(\theta_{K}^t\right)\right\Vert^2\leq& \frac{8\left(f_{1:K}^0 - f_{1:K}^*\right)}{\eta_G\eta_LEKT} + {4\eta_G\eta_LEL\sigma_L^2} + \frac{4\left(M-N\right)}{N\left(M-1\right)}\eta_G\eta_L{ \beta}EL\sigma_G^2\allowdisplaybreaks\nonumber\\
        & + 80\left[1 + \Lambda_{1:K-1} + {}\Lambda_{K} \right]\eta_L^2E^2D^2L^2 { + 4\beta \sigma_G^2},
    \end{align}
    where $f_{1:K}^0\triangleq f_{1:K}\left(\theta_K^0\right)$, $f_{1:K}^*\triangleq f_{1:K}\left(\theta_K^*\right)$, $\Lambda_K\triangleq \frac{1}{T}\sum_{t=1}^{T}\mathbb{E}_t\left\Vert \frac{1}{EN}\sum_{m\in\mathcal{S}_K^t}\sum_{e=0}^{E-1}w_{K,m}^{t,e,*}\right\Vert^2$, and $\Lambda_{1:K-1} \triangleq \frac{1}{T}\sum_{t=1}^{T}\mathbb{E}_t\left\Vert \frac{1}{EN}\sum_{m\in\mathcal{S}_K^t}\sum_{e=0}^{E-1}\left(1-w_{K,m}^{t,e,*}\right)\right\Vert^2$.
\end{theorem}

{
\textbf{Implications.}
Eq.~(\ref{eq:weighted convergence}) separates optimization progress from the effects of stochasticity, client sampling, and local updates. The first term depends on the initial optimality gap $f_{1:K}^0-f_{1:K}^*$ and decreases with $T$ for fixed admissible learning rates and local-step count $E$. The second term captures local stochastic variance $\sigma_L^2$, while the third captures the contribution of client heterogeneity under partial participation. Its factor $(M-N)/(N(M-1))$ decreases as more clients participate and vanishes when $N=M$.

The local-update term scales with $\eta_L^2E^2D^2L^2$ and depends on the weight moments $\Lambda_{1:K-1}$ and $\Lambda_K$. These quantities summarize the total weight assigned to historical tasks and to the current task, respectively. Since $w_K^{t,*}\in[0,1]$, their sum lies in $[1/2,1]$. Thus, adaptive weight allocation changes the coefficient of this term within a bounded range. Choosing $\eta_L$ proportional to $1/E$ controls its explicit growth with the number of local steps; increasing $E$ alone does not necessarily tighten the overall bound.
The remaining term $4\beta\sigma_G^2$ reflects the global-local gradient discrepancy allowed by Assumption~\ref{assumption5:Bounded Global Gradient Variance}. Unlike the sampling-related heterogeneity term, it persists under full participation and is not reduced by the learning rate alone. For fixed $K$, $\eta_G$, and problem parameters, choosing $\eta_L=\Theta(1/(EL\sqrt{T}))$ reduces the other terms at rates $O(T^{-1/2})$ or faster, leaving a stationarity bound of $4\beta\sigma_G^2+O(T^{-1/2})$. The result therefore quantifies how optimization and local training interact with the heterogeneity tolerance in the bound.}

\begin{remark}[Relation to prior convergence analyses]
{\citet{han2023convergence} and \citet{keshri2025convergence} analyze adaptive updates that balance a pooled historical objective with the current objective, including stationarity guarantees for the historical objective. Here, historical tasks retain separate weights, and the quantity $\nabla\tilde f_{1:K}(\theta_K^t)$ tracks the federation-wide gradient combination induced by the client-specific local weighting decisions.
\citet{xu2026serverproximal} establish a convergence bound for the cumulative objective $f_{1:K}$ under a bounded inter-task gradient difference assumption. Its bound explicitly includes the initial historical-gradient norm $\|\nabla f_{1:K-1}(\theta_K^0)\|^2$.  Eq.~(\ref{eq:weighted convergence}) instead evaluates stationarity under adaptive task weights and makes their historical--current allocation explicit through $\Lambda_{1:K-1}$ and $\Lambda_K$. The two results concern different stationarity criteria. The distinguishing feature here is the analysis of a global task-gradient combination induced by local adaptive decisions, with separate terms for partial participation and persistent global-local gradient discrepancy.}
\end{remark}

\begin{corollary}[Convergence Rate]\label{corollary:Convergence Rate}
    With a constant learning rate $\eta_G$ and a local learning rate $\eta_L = \mathcal{O}(\frac{1}{\sqrt{KT}EL})$, and if $\beta = \mathcal{O}(\eta_L)$, when the length of the task sequence $K$ is fixed, the weighted task-uniform convergence rate satisfies
    \begin{equation}
        \frac{1}{T}\sum_{t=0}^{T-1}\mathbb{E}_t\left\Vert\nabla \tilde{f}_{1:K}\left(\theta_{K}^t\right)\right\Vert^2 = \mathcal{O}\left(\frac{1}{\sqrt{T}} + \frac{\sigma_L^2}{\sqrt{T}} + \frac{\sigma_G^2}{E\sqrt{T}}\right).
    \end{equation}
\end{corollary}

{
Corollary~\ref{corollary:Convergence Rate} makes explicit the role of persistent client heterogeneity. For a fixed nonzero $\beta$, \Adaptive converges to a heterogeneity-dependent stationarity neighborhood, with the remaining optimization and stochastic terms decaying as $\mathcal{O}(T^{-1/2})$ or faster. Under the additional condition $\beta=\mathcal{O}(\eta_L)$, this residual also vanishes, yielding an overall $\mathcal{O}(T^{-1/2})$ weighted task-uniform stationarity rate. This has the same dependence on the number of communication rounds as nonconvex federated multi-objective optimization guarantees \citep{yang2023federated, askin2025federated}, although those results consider simultaneous access to the objectives and use different stationarity criteria.}

\section{Experiments}
\label{sec:experiments}

We evaluate \Adaptive on class- and domain-incremental benchmarks and examine its performance and the contributions of shared historical-gradient reconstruction and adaptive task weighting. All experiments are implemented in PyTorch with two NVIDIA Quadro RTX 8000 GPUs. Additional sensitivity studies of task weights, curvature approximations, and runtime are {provided in Appendix~\ref{sec:details of experiments}.}

\textbf{Setup.}
We use CIFAR-100~\citep{krizhevsky2009learning} and ImageNet100~\citep{russakovsky2015imagenet} for class-incremental learning, each divided into five tasks of 20 classes. For domain-incremental learning, Digit-10 and VLCS~\citep{torralba2011unbiased} each comprise four sequential domain tasks. Unless otherwise stated, we use $M=16$ clients, $N=8$ participants per round, and $T=50$ communication rounds per task. Each participating client performs three {local gradient steps} per round. Client data follow Dirichlet partitions with $\alpha=1$ for the class-incremental benchmarks and $\alpha=0.5$ for the domain-incremental benchmarks. We use ResNet-18~\citep{he2016deep}, batch size 32, learning rates $\eta_L=0.001$ and $\eta_G=1$, and a private replay buffer of 400 examples per client. Main results are averaged over three seeds, and component and sensitivity studies use CIFAR-100. Dataset construction, implementation details, and baseline adaptations {are described in Appendix~\ref{sec:detailed configurations}.}

\textbf{Baselines.}
We compare with conventional FL methods FedAvg~\citep{McMahan2017FedAvg} and FedProx~\citep{li2020federated}; multi-objective FL methods FSMGDA~\citep{yang2023federated} and FedCMOO~\citep{askin2025federated}, adapted to sequential training with replay; and FCL methods FLwF-2T~\citep{usmanova2022federated}, FedSSI~\citep{li2025fedssi}, FedAGC~\citep{zhang2025fedagc}, SPECIAL~\citep{xu2026serverproximal}, and C-FLAG~\citep{keshri2025convergence}. We also evaluate our federated adaptations of POCL~\citep{wu2024mitigating} and BECAME~\citep{li2025became}, denoted FedPOCL and FedBECAME.

\textbf{Metrics.}
Let $A_{i,j}$ denote test accuracy on task $j$ after learning task $i$. We report final average accuracy (ACC) and backward transfer (BWT):
\begin{equation}
\mathrm{ACC}=\frac{1}{K}\sum_{j=1}^{K}A_{K,j},
\qquad
\mathrm{BWT}=\frac{1}{K-1}\sum_{j=1}^{K-1}(A_{K,j}-A_{j,j}).
\label{eq:experimental-metrics}
\end{equation}
Higher values are better for both metrics, and negative BWT indicates an average decline on earlier tasks relative to their performance immediately after learning. ACC is reported in percent, while BWT and differences between reported accuracies are measured in percentage points (pp).

\begin{table*}[t]
\centering
\caption{Final average accuracy (ACC, \%) and backward transfer (BWT, pp) on four benchmarks. Results are averaged over three seeds. Bold denotes the best reported mean in each column.}
\vspace{-0.1in}
\label{tab:main_results}
\resizebox{\textwidth}{!}{%
\begin{tabular}{lcccccccc}
\toprule
& \multicolumn{2}{c}{CIFAR-100}
& \multicolumn{2}{c}{ImageNet100}
& \multicolumn{2}{c}{Digit-10}
& \multicolumn{2}{c}{VLCS} \\
\cmidrule(lr){2-3}\cmidrule(lr){4-5}\cmidrule(lr){6-7}\cmidrule(lr){8-9}
Method & ACC (\%) $\uparrow$ & BWT (pp) $\uparrow$
& ACC (\%) $\uparrow$ & BWT (pp) $\uparrow$
& ACC (\%) $\uparrow$ & BWT (pp) $\uparrow$
& ACC (\%) $\uparrow$ & BWT (pp) $\uparrow$ \\
\midrule
FedAvg & $43.21\pm1.07$ & $-19.09\pm1.87$
& $28.94\pm0.66$ & $-25.65\pm0.66$
& $61.83\pm1.22$ & $-43.72\pm0.98$
& $53.33\pm0.74$ & $-18.81\pm0.21$ \\
FedProx & $43.25\pm2.42$ & $-20.92\pm2.21$
& $31.55\pm1.83$ & $-24.01\pm2.08$
& $60.55\pm1.40$ & $-45.68\pm0.73$
& $44.99\pm0.32$ & $-22.69\pm0.88$ \\
\midrule
FSMGDA & $34.00\pm0.52$ & $-11.75\pm0.80$
& $21.93\pm2.01$ & $-14.35\pm3.89$
& $67.31\pm0.84$ & $-21.14\pm0.44$
& $53.78\pm2.29$ & $-5.01\pm1.80$ \\
FedCMOO & $45.09\pm2.93$ & $-11.37\pm1.64$
& $31.04\pm2.12$ & $-11.82\pm2.59$
& $69.46\pm0.80$ & $-20.01\pm0.41$
& $54.79\pm0.61$ & $-4.26\pm0.85$ \\
\midrule
FLwF-2T & $34.01\pm1.21$ & $-17.66\pm2.31$
& $29.14\pm1.26$ & $-12.49\pm2.33$
& $68.31\pm0.70$ & $-21.00\pm0.35$
& $51.07\pm0.29$ & $-7.69\pm0.49$ \\
FedSSI & $44.77\pm0.55$ & $-10.07\pm0.09$
& $32.52\pm0.50$ & $-10.38\pm0.59$
& $52.37\pm1.45$ & $-17.80\pm2.14$
& $49.22\pm3.20$ & $-9.16\pm1.91$ \\
SPECIAL & $43.17\pm1.69$ & $-10.07\pm1.61$
& $31.97\pm0.68$ & $-13.98\pm1.04$
& $66.01\pm2.31$ & $-22.09\pm3.13$
& $53.61\pm0.98$ & $-5.04\pm0.37$ \\
FedPOCL & $47.09\pm2.29$ & $-10.17\pm1.43$
& $42.01\pm2.40$ & $-12.42\pm1.40$
& $75.00\pm2.51$ & $\mathbf{-16.47\pm0.87}$
& $56.31\pm3.09$ & $-5.07\pm1.19$ \\
FedAGC & $41.81\pm0.73$ & $-12.78\pm0.29$
& $33.81\pm3.25$ & $-10.99\pm2.51$
& $62.71\pm2.90$ & $-21.49\pm2.20$
& $52.74\pm1.88$ & $-3.39\pm0.78$ \\
FedBECAME & $42.13\pm1.77$ & $-11.98\pm2.37$
& $30.90\pm2.08$ & $-21.87\pm2.08$
& $63.89\pm1.90$ & $-22.37\pm4.31$
& $51.92\pm3.21$ & $-5.09\pm1.94$ \\
C-FLAG & $46.82\pm2.71$ & $-13.05\pm2.73$
& $36.75\pm1.83$ & $-11.99\pm2.32$
& $69.21\pm1.36$ & $-19.80\pm2.44$
& $53.10\pm1.97$ & $-3.92\pm1.38$ \\
\midrule
\Adaptive (ours) & $\mathbf{54.66\pm0.15}$ & $\mathbf{-6.07\pm0.29}$
& $\mathbf{43.31\pm0.89}$ & $\mathbf{-6.92\pm0.93}$
& $\mathbf{80.12\pm0.93}$ & $-18.56\pm0.68$
& $\mathbf{57.03\pm0.46}$ & $\mathbf{-2.28\pm0.23}$ \\
\bottomrule
\end{tabular}}
\vspace{-0.05in}
\end{table*}

\textbf{Main results.}
Table~\ref{tab:main_results} shows that \Adaptive achieves the highest mean final accuracy on all four benchmarks. Compared with FedPOCL, the strongest baseline by mean ACC on each dataset, the gains are 7.57 and 1.30~pp on CIFAR-100 and ImageNet100, and 5.12 and 0.72~pp on Digit-10 and VLCS. The improvements extend across both incremental settings, with the largest gains on CIFAR-100 and Digit-10.
\Adaptive also achieves the highest BWT on both class-incremental benchmarks and VLCS. On Digit-10, however, FedPOCL has better BWT ($-16.47$~pp vs. $-18.56$~pp), despite its lower final accuracy ($75.00\%$ vs. $80.12\%$). These metrics capture complementary aspects of continual learning: \textit{final accuracy measures performance after learning the task sequence, whereas BWT measures changes relative to each method's own performance immediately after learning an earlier task.}
Both methods adapt task weights, but our federated POCL implementation uses local replay, whereas \Adaptive uses shared reconstructed history. 
Table~\ref{tab:component_ablation} further examines the contributions of the historical-gradient source and adaptive weighting.

\begin{table}[t]
\centering
\caption{Historical-gradient source and weighting on CIFAR-100. Local replay uses each client's retained data at its current model; shared reconstruction uses the aggregated gradients and curvature summaries of \Adaptive. Fixed weights are shared across clients, whereas adaptive weights are recomputed during local training. The final row is the full \Adaptive method. }
\vspace{-0.1in}
\label{tab:component_ablation}
\small
\begin{tabular}{llcc}
\toprule
Historical gradients & Weighting & ACC (\%) $\uparrow$ & BWT (pp) $\uparrow$ \\
\midrule
{Shared w/o curvature} & {Adaptive} & {$40.19$} & {$-13.94$} \\
Local replay & Fixed & $35.27$ & $-17.18$ \\
Local replay & Adaptive & $47.57$ & $-12.69$ \\
Shared reconstruction & Fixed & $39.54$ & $-11.84$ \\
Shared reconstruction & Adaptive & $\textbf{54.66}$ & $-\textbf{6.07}$ \\
\bottomrule
\end{tabular}
\vspace{-0.15in}
\end{table}

\begin{wrapfigure}{r}{.5\linewidth}
    \centering
    \includegraphics[width=1\linewidth]{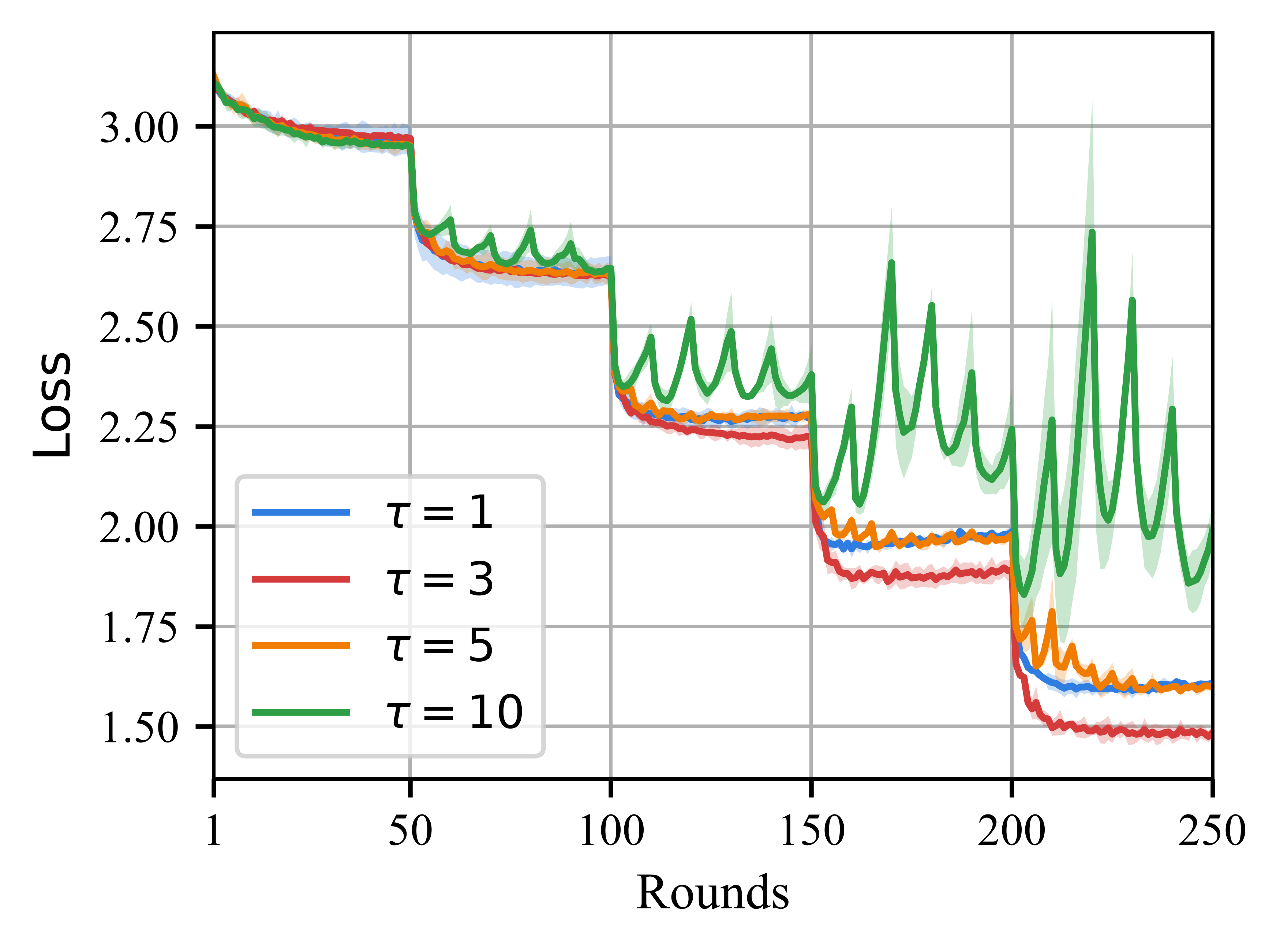} 
    \vspace{-0.25in}
    \caption{Loss of \Adaptive on CIFAR-100 for curvature-refresh interval $\tau$. Historical-task curvature summaries are refreshed every $\tau$ communication rounds. }
    \label{fig:frequency}
    \vspace{-0.15in}
\end{wrapfigure}
\textbf{Historical-gradient source and adaptive weighting.}
Table~\ref{tab:component_ablation} compares fixed and adaptive weights using either local replay gradients or shared reconstructed gradients. Adaptive weighting improves ACC by 12.30~pp with local replay and by 15.12~pp with shared reconstruction. Conversely, replacing local replay with shared reconstruction improves ACC by 4.27~pp under fixed weights and by 7.09~pp under adaptive weights. { The curvature information improves ACC by 14.47~pp by refining the reconstructed gradients of previous tasks.} \textit{Both components therefore contribute, and the gain from shared reconstruction is larger when weights are adapted.}

The full method also achieves the highest BWT among these variants, improving it by 6.62~pp over adaptive weighting with local replay. Together with the corresponding 7.09~pp ACC gain, this result shows that \textit{adaptive weighting benefits from using shared  reconstructed gradients rather than relying on each client's buffer.}

\begin{wrapfigure}{r}{.5\linewidth}
    \centering
    \includegraphics[width=1\linewidth]{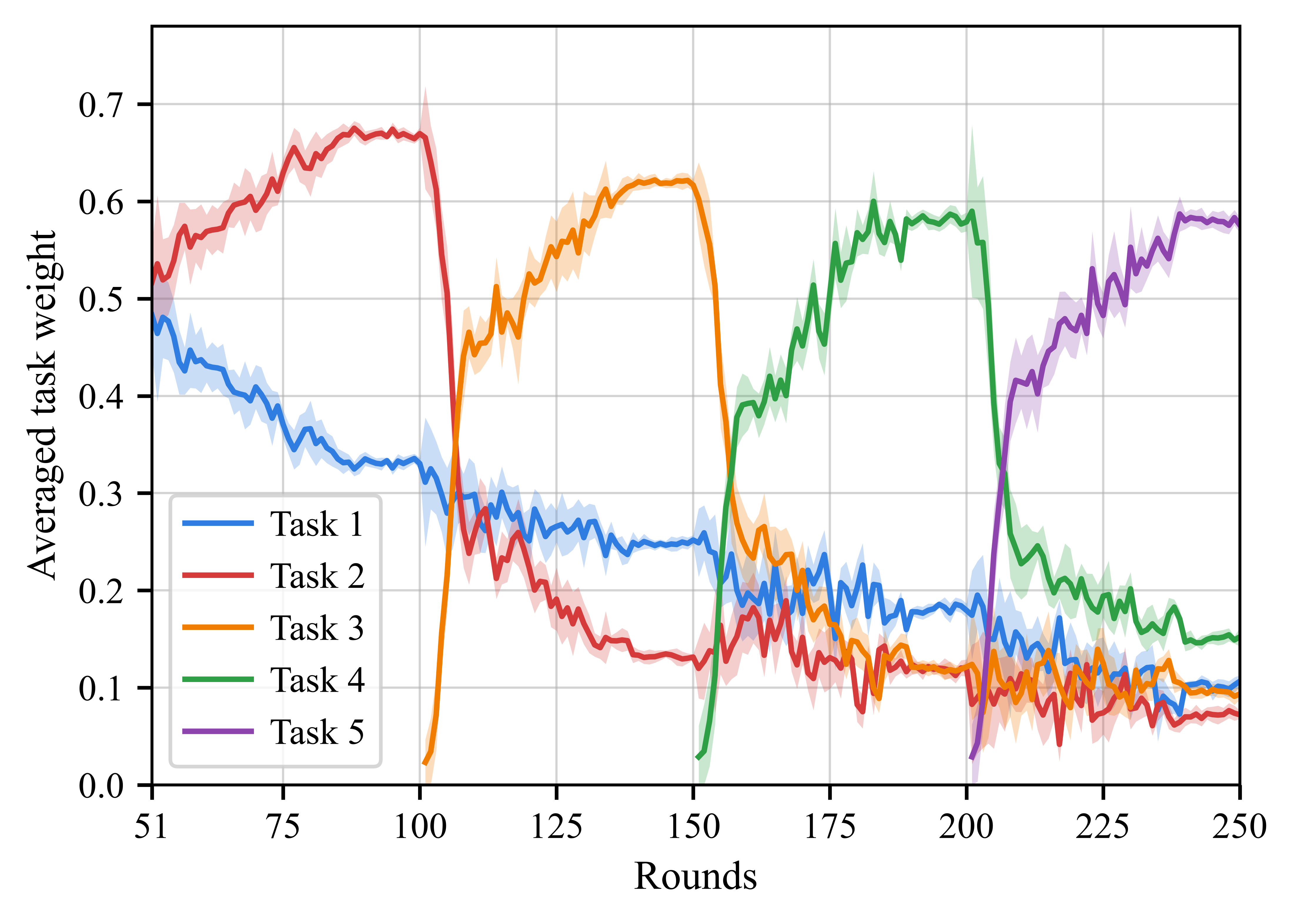} 
    \vspace{-0.25in}
    \caption{Evolution of averaged task weights in \Adaptive on CIFAR-100 over communication rounds. }
    \label{fig:task weight}
    \vspace{-0.2in}
\end{wrapfigure}
\textbf{Curvature refresh frequency.} { Figure~\ref{fig:frequency} shows pronounced loss oscillations when the refresh interval reaches $10$ rounds, while shorter intervals are more stable. This illustrates the sensitivity to outdated curvature information, and evaluating the overall efficiency trade-off additionally requires communication and runtime measurements.}{ Additionally, the curvature rank $r$ and replay buffer size affect learning performance and resource consumption. The results show that rank $3$ and a replay budget of $400$ samples per client provide a favorable balance between accuracy, knowledge retention, and computational overhead. Detailed results and analyses are provided in Appendix~\ref{sec:Additional Ablation Studies}.}

\textbf{Adaptive task weights.}
{Figure~\ref{fig:task weight} illustrates how \Adaptive reallocates task weights during sequential training. Within the displayed stages, the current-task weight generally increases and exceeds the individual historical-task weights later in training. Its weight drops when the next task arrives, while historical-task weights continue to evolve without a fixed ordering. These trajectories illustrate adaptive allocation both between current and historical tasks and among historical tasks.}

\clearpage

\bibliography{iclr2027_conference}
\bibliographystyle{iclr2027_conference}

\clearpage
\appendix
\section{Related Work}
\subsection{Continual Learning}
Continual Learning aims to enable a model to sequentially learn knowledge from evolving tasks while mitigating catastrophic forgetting of previously learned knowledge. Existing approaches can be categorized into replay-based, regularization-based, and architecture-based methods. Replay-based methods maintain a limited memory of selected previous samples and jointly incorporate them with the current data during training~\citep{chaudhry2019tinyepisodicmemoriescontinual, Rebuffi2017iCaRL, wu2018memory}. Regularization-based methods preserve previous predictions through knowledge distillation and prevent the model from deviating from the previous optimal model~\citep{Kirkpatrick2017EWC, LiHoiem2017LwF, ritter2018online, li2025became}. Architecture-based methods mitigate catastrophic forgetting by expanding the network or task-specific parameters to isolate the information learned from different tasks~\citep{mallya2018packnet, Liang2023Adaptive}. 

\subsection{Federated Continual Learning}
Federated Continual Learning can be regarded as an expansion of conventional continual learning in the federated setting, which enables clients to sequentially train the model cooperatively, so the main types of FCL can also be categorized into these three types~\citep{yoon2021federated, usmanova2022federated, babakniya2023data, wuerkaixi2025accurate}. However, a key distinction between the two settings lies in information communication, as the communication between clients and the server in FCL needs to account for limited bandwidth and communication costs~\citep{wu2021fast, zhang2023communication, li2024towards, paul2023masked}. In addition, FCL introduces task-level asynchrony besides the communication asynchrony inherited from conventional FL, where different clients may simultaneously train on different tasks due to the heterogeneous task sequences of different clients. FedSpace addresses this asynchrony through prototype aggregation, representation alignment and a modified aggregation policy, enabling knowledge sharing among different clients~\citep{shenaj2023AFCL}, and C$^2$-AFCL introduces a two-stage orthogonal calibration mechanism to mitigate both client and task drift caused by asynchronous client updates~\citep{li2026crosstask}.
{
HERO~\citep{nguyen2026hero} distinguishes client data skew from client task-order mismatch, enabling these dimensions of FCL heterogeneity to be evaluated separately. \Adaptive focuses on the former under a shared task sequence: clients contribute different replay-gradient and curvature estimates for the same historical task, which are aggregated to support adaptive local training. Extending this protocol to client-specific task orders requires accounting for differences in task availability and current versus historical status across clients.}

\subsection{Hessian-informed Continual Learning}
Second-order information has been widely explored in CL to preserve previously acquired knowledge. For all Hessian-based methods, the problem which can not be overlooked is how to find the Hessian matrix. Since forming the full Hessian causes unnecessary computation and storage costs, recent studies focus on using more efficient approximation approaches to recover the Hessian information while training.  The Fisher information matrix is one of the most widely used approximation methods, whose diagonal entries are commonly employed to measure the importance of parameters, but it only captures parameter-wise information while ignoring the correlations among parameters, leading to substantial approximation errors~\citep{Kirkpatrick2017EWC, shoham2019overcomingforgettingfederatedlearning, liu2022continual}. As an alternative to the Fisher information matrix, ~\citet{daxberger2023improving} use the generalized Gauss–Newton (GGN) matrix to provide a more accurate curvature approximation for previous-gradient reconstruction. Another challenge of utilizing Hessian information in CL setting is to prevent the model parameters from deviating significantly from the point at which the Hessian was computed. ~\citet{wu2024meta} and ~\citet{li2026converge} employ iterative strategies to update the Hessian to improve the accuracy of approximation by maintaining up-to-date curvature information while approximating. However, effectively maintaining and updating Hessian information in distributed learning settings remains largely underexplored.

\section{Notation List}\label{sec:notation}
The detailed notation is listed in table

\begin{table}[ht]
    \centering
    \caption{ and Terminologies.}
    \begin{tabular}{|c|c|}
    \hline
    \textbf{Notation} & \textbf{Description} \\
    \hline
    $M$ & Number of all clients \\
    \hline
    $N$ & Number of participating clients per round\\
    \hline
    $T$ & Number of global rounds per task \\
    \hline
    $E$ & Number of local gradient steps \\
    \hline
    $K$ & Number of tasks \\
    \hline
    $[n]$ & Set of integers $\left\{1,\ldots,n\right\}$\\
    \hline
    $\theta_i^t$ & Global model at round $t$ in task $i$\\
    \hline
    $\theta_{i,m}^t$ & Local model on client $m$ at round $t$ in task $i$\\
    \hline
    $\theta_{i,m}^{t,e}$ & Local model on client $m$ at epoch $e$ of round $t$ in task $i$\\
    \hline
    $w_{i,m}^{t,e,*}$ & Tuned weight for the gradient of task $i$ on client $m$ of epoch $e$ on round $t$\\ 
    \hline
    $\boldsymbol{\mathrm{d}}_{i,m}^{t,e}$ & Update direction on client $m$ of epoch $e$ on round $t$\\ \hline
    $\Delta_{i,m}^t$ & Update vector on client $m$ of round $t$ in task $i$\\
    \hline
    $f_{i}$ & Loss function of task $i$\\
    \hline
    $f_{1:K}$ & Loss function over task $1$ to task $K$\\
    \hline
    $\nabla f_{i}$ & the gradient of $f_i$\\
    \hline
    $\nabla \tilde{f}_{1:K}$ & Weighted gradient over all tasks\\
    \hline
    $\mathcal{T}$ & Set of all tasks, i.e., $\{\mathcal{T}_1,\ldots, \mathcal{T}_K\}$ \\
    \hline
    $\mathcal{T}_i$ & Dataset of the task $i$\\
    \hline
    $\mathcal{T}_{i,m}$ & Dataset of client $m$ on task $i$ \\
    \hline
    $\mathcal{S}_{i}^t$ & The index of participants at round $t$ in task $i$ \\ \hline
    $\boldsymbol{w}_{m}^{t,e,*}$ & Set of tuned weights on client $m$ of epoch $e$ on round $t$ \\ \hline
    $\left\Vert\cdot\right\Vert^2$ & $\ell_2$-norms\\
    \hline
    \end{tabular}
    \label{Table: notation}
    \end{table}

\section{Algorithms}
\subsection{Lanczos Algorithm}\label{sec:lanczos algorithm}
Lanczos Algorithm is an iterative method for efficiently approximating the dominant eigenvalues and eigenvectors of a large symmetric matrix. A prerequisite for a matrix to be decomposed via Lanczos algorithm is the existence of an operator $\mathcal{H}(\boldsymbol{q}):\boldsymbol{q}\mapsto
\boldsymbol{H}\boldsymbol{q}$, and the automatic differentiation for $\left.\nabla_\theta[\nabla_\theta\widehat f_{i,m}(\theta)^\top\boldsymbol q]\right|_{\theta=\theta_K^t}$ enables the product of the Hessian matrix and any vector $\boldsymbol{q}\in\mathbb{R}^d$ to be computed, which guarantees the original Hessian matrix to be decomposed via Lanczos Algorithm. The full procedure is listed in Algorithm~\ref{alg:Hessian-Lanczos Decomposition}.

\begin{algorithm}[h]
\caption{Hessian-Lanczos Decomposition Algorithm}
\label{alg:Hessian-Lanczos Decomposition}
\begin{algorithmic}[1]
\item[] \hspace*{-1.26em}\textbf{Input:} Operator 
$\mathcal{H}(\boldsymbol{q}):\boldsymbol{q}\mapsto
\boldsymbol{H}\boldsymbol{q}$, target rank $r$, Lanczos steps $k\geq r$

\item[] \hspace*{-1.6em} \textbf{Initialize:} 
$\boldsymbol{q}_1\in\mathbb{R}^{d\times 1}$ with $\|\boldsymbol{q}_1\|=1$,
$\boldsymbol{q}_0=\boldsymbol{0}\in\mathbb{R}^{d\times 1}$, $\beta_0=0$

\FOR{$j=1,\ldots,k$}
    \STATE $\boldsymbol{w}\leftarrow
    \boldsymbol{H}\boldsymbol{q}_j-\beta_{j-1}\boldsymbol{q}_{j-1}$
    \STATE $\alpha_j\leftarrow\boldsymbol{q}_j^T\boldsymbol{w}$
    \STATE $\boldsymbol{w}\leftarrow
    \boldsymbol{w}-\alpha_j\boldsymbol{q}_j$
    \STATE Reorthogonalize $\boldsymbol{w}$ against
    $\{\boldsymbol{q}_1,\ldots,\boldsymbol{q}_j\}$
    \STATE $\beta_j\leftarrow\|\boldsymbol{w}\|$,
    $\boldsymbol{q}_{j+1}\leftarrow\boldsymbol{w}/\beta_j$
\ENDFOR
\STATE Construct $\boldsymbol{Q}_k=[\boldsymbol{q}_1,\ldots,\boldsymbol{q}_k]$
and tridiagonal $\boldsymbol{T}_k$ from $\{\alpha_j,\beta_j\}$
\STATE Decompose $\boldsymbol{T}_k$ as $\boldsymbol{T}_k=\boldsymbol{V}\boldsymbol{\Theta}\boldsymbol{V}^T$
\STATE Select the top-$r$ eigenpairs $(\boldsymbol{V}_r,\boldsymbol{\Theta}_r)$
\STATE Reconstruct $\widetilde{\boldsymbol{Q}}\leftarrow\boldsymbol{Q}_k\boldsymbol{V}_r$,
$\widetilde{\boldsymbol{\Lambda}}\leftarrow\boldsymbol{\Theta}_r$
\RETURN $\widetilde{\boldsymbol{Q}}_{d\times r},\widetilde{\boldsymbol{\Lambda}}_{r\times r}$\\

\end{algorithmic}
\end{algorithm}

\subsection{Frank-Wolfe Algorithm}\label{sec:frank wlfe algorithm}
In order for the Frank-Wolfe Algorithm to be self-contained, we restate the algorithm in \citet{wu2024mitigating} here as Algorithm~\ref{alg:frank wolfe}.
\begin{algorithm}[h]
\caption{Frank-Wolfe Algorithm for Solving Eq. (\ref{eq:adaptive task weights})}
\label{alg:frank wolfe}
\begin{algorithmic}[1]
\item[] \hspace*{-1.26em}\textbf{Input:} The gradients set $\boldsymbol{G} = [\nabla\tilde{f}_1(\theta_{K,m}^{t,e})^\top, \ldots, \nabla\tilde{f}_{K-1}(\theta_{K,m}^{t,e})^\top, \nabla f_{K,m}(\theta_{K,m}^{t,e};\xi_{K,m}^{t,e})^\top]$

\item[] \hspace*{-1.6em} \textbf{Initialize:} 
$\boldsymbol{w} = [\frac{1}{K}, \ldots, \frac{1}{K}]$

\STATE Compute $\boldsymbol{P} = \boldsymbol{G}^\top \boldsymbol{G}$
\REPEAT
    \STATE $\boldsymbol{\alpha} = \arg\min_{\boldsymbol{\alpha}\in\{\boldsymbol{\alpha}^\top \boldsymbol{1} = 1, \boldsymbol{\alpha}\succ0\}}\boldsymbol{\alpha}^\top\boldsymbol{P}\boldsymbol{w}$
    \STATE $L = \boldsymbol{w} + \eta(\boldsymbol{\alpha} - \boldsymbol{w})$
    \STATE $\eta = \arg\min_{\eta\in[0,1]} L^\top \boldsymbol{P}L$
    \STATE $\boldsymbol{w} \leftarrow (1-\eta)\boldsymbol{w}+\eta\boldsymbol{\alpha}$

\UNTIL $\eta\sim 0$ or reaching the maximum iteration number
\RETURN The final computed weights $\boldsymbol{w}$\\

\end{algorithmic}
\end{algorithm}

\subsection{Distributed Spectral Hessian Aggregation Algorithm}\label{sec:Distributed Spectral Hessian Aggregation Algorithm}

We list the algorithm of DSHA in Algorithm~\ref{alg:Centralized Low-rank EVD}.

\begin{algorithm}[h]
\caption{Distributed Spectral Hessian Aggregation (DSHA)}
\label{alg:Centralized Low-rank EVD}
\begin{algorithmic}[1]
\REQUIRE $\boldsymbol{A}_{K,i}^t$, $\boldsymbol{B}_{K,i}^t$, target rank $r$
\STATE Compute an economy QR factorization $\boldsymbol{A}_{K,i}^t=\boldsymbol{D}_{K,i}^t\boldsymbol{R}_{K,i}^t$.
\STATE Form $\boldsymbol{C}_{K,i}^t= \boldsymbol{R}_{K,i}^t\boldsymbol{B}_{K,i}^t(\boldsymbol{R}_{K,i}^t)^\top$.
\STATE Retain at most $r$ eigenpairs of $\boldsymbol{C}_{K,i}^t$ with largest eigenvalue magnitudes in $(\boldsymbol{V}_{K,i}^t,\boldsymbol{\Gamma}_{K,i}^t)$, preserving the eigenvalue signs.
\STATE Set $\boldsymbol{U}_{K,i}^t=\boldsymbol{D}_{K,i}^t\boldsymbol{V}_{K,i}^t$.
\STATE \textbf{return} $(\boldsymbol{U}_{K,i}^t,\boldsymbol{\Gamma}_{K,i}^t)$.
\end{algorithmic}
\end{algorithm}

\section{Lemmas}
\begin{lemma}\label{lemma:Hessian}
    Suppose $f$ is twice differentiable and $L$-smooth, let $\boldsymbol{H} = \nabla^2 f(x)$, $\theta\in\mathbb{R}^d$, and $\tilde{\boldsymbol{H}} = \boldsymbol{U}\boldsymbol{V}\boldsymbol{U}^T$ as the rank-$r$ truncated eigendecomposition of $\boldsymbol{H}$, we have:
    \begin{equation}
        \left\Vert \tilde{\boldsymbol{H}}\theta\right\Vert^2\leq L^2\left\Vert\theta\right\Vert^2.
    \end{equation}
\end{lemma}

\begin{proof}
    Since $f$ is $L$-smooth and twice differentiable, then we have
    \begin{equation}
        \left\Vert\boldsymbol{H}\right\Vert_2 = \left\Vert\nabla^2 f\left(x\right)\right\Vert_2\leq L.
    \end{equation}
    Since $\boldsymbol{H}$ is symmetric, its spectral norm is given by the largest absolute eigenvalue as
    \begin{equation}
        \left\Vert\boldsymbol{H}\right\Vert_2 = \max_{i\in[d]}|\lambda_i(\boldsymbol{H})|.
    \end{equation}
    Therefore, every eigenvalue $\lambda_i, i\in[r]$, retained in the diagonal elements of $\boldsymbol{V}$ satisfies
    \begin{equation}
        |\lambda_i|\leq L,
    \end{equation}
    which means that $\left\Vert\boldsymbol{V}\right\Vert_2 = \max_{i\in[r]}|\lambda_i|\leq L$.
    Since $\boldsymbol{U}$ is a matrix with orthonormal columns, its columns are orthonormal, then we have $\boldsymbol{U}^\top\boldsymbol{U} = \boldsymbol{I}_r$, which means $\left\Vert\boldsymbol{U}\right\Vert_2 = \left\Vert\boldsymbol{U}^T\right\Vert_2 = 1$. Therefore, for any $\theta\in\mathbb{R}^d$, we have 
    \begin{align}
        \left\Vert \tilde{\boldsymbol{H}}\theta\right\Vert =& \left\Vert \boldsymbol{U}\boldsymbol{V}\boldsymbol{U}^T\theta\right\Vert\allowdisplaybreaks\nonumber\\
        \leq &\Vert \boldsymbol{U}\Vert_2 \Vert\boldsymbol{V}\Vert_2 \Vert\boldsymbol{U}^T\Vert_2 \Vert\theta\Vert\allowdisplaybreaks\nonumber\\
        \leq& L\Vert\theta\Vert,
    \end{align}
    and we can further have $\Vert \tilde{\boldsymbol{H}}\theta\Vert^2 \leq L^2\Vert\theta\Vert^2$, which completes the proof.

\end{proof}

\begin{lemma}\label{lemma:Hessian2}
    The curvature matrix $\tilde{\boldsymbol{H}}_{K,i}^t$ used by clients to reconstruct the gradients satisfies that $\Vert \tilde{\boldsymbol{H}}_{K,i}^t\theta\Vert^2\leq L^2\Vert\theta\Vert^2, \forall \theta\in\mathbb{R}^d$.
\end{lemma}
\begin{proof}
    According to Lemma \ref{lemma:Hessian}, the updated curvature information $(\tilde{\boldsymbol{Q}}_{K,i,m}^t, \tilde{\boldsymbol{\Lambda}}_{K,i,m}^t)$ satisfies $\Vert\tilde{\boldsymbol{Q}}_{K,i,m}^t\tilde{\boldsymbol{\Lambda}}_{K,i,m}^t(\tilde{\boldsymbol{Q}}_{K,i,m}^t)^\top\Vert^2\leq L^2$, then we have
    \begin{align}
        \Vert \bar{\boldsymbol{H}}_{K,i}^t\Vert^2 = \left\Vert\frac{1}{N}\sum_{m\in\mathcal{S}^t_K}\tilde{\boldsymbol{Q}}_{K,i,m}^t\tilde{\boldsymbol{\Lambda}}_{K,i,m}^t(\tilde{\boldsymbol{Q}}_{K,i,m}^t)^\top\right\Vert^2\leq L^2,
    \end{align}
    which means that the $\ell^2$-norm of recovered Hessian matrix is still bounded by $L^2$. Therefore, according to Lemma \ref{lemma:Hessian}, we also have 
    \begin{equation}
        \left\Vert\tilde{\boldsymbol{H}}_{K,i}^t\theta\right\Vert^2 = \left\Vert{\boldsymbol{U}}_{K,i}^t{\boldsymbol{\Gamma}}_{K,i}^t({\boldsymbol{U}}_{K,i}^t)^\top\theta\right\Vert^2\leq L^2\Vert\theta\Vert^2, 
    \end{equation}
    where $({\boldsymbol{U}}_{K,i}^t, {\boldsymbol{\Gamma}}_{K,i}^t)$ are the results of truncated EVD of ${\boldsymbol{H}}_{K,i}^t$ with rank-$r$.
\end{proof}

\begin{lemma}\label{lemma:bounded local updates}
If the local learning rate satisfies $\eta_L\leq \frac{1}{3(k-1)EL}, \forall k\geq 2$, we have the following results:
    \begin{equation}
        \left\Vert\theta_{k,m}^{t,e} - \theta_k^t \right\Vert^2\leq20\eta_L^2E^2D^2.
    \end{equation}
\end{lemma}

\begin{proof}
    \begin{align}
        &\left\Vert\theta_{k,m}^{t,e} - \theta_K^t \right\Vert^2\allowdisplaybreaks\nonumber\\
        =& \left\Vert\theta_{k,m}^{t,e-1} - \theta_K^t - \eta_L\left(\sum_{i=1}^{k-1}w_{i,m}^{t,e-1,*} \left(\nabla f_{i}(\theta_{k}^t) + \tilde{\boldsymbol{H}}_i\left(\theta_{k,m}^{t,e-1} - \theta_K^t\right)\right) \right.\right.\allowdisplaybreaks\nonumber\\
        &\left.\left.+w_{K,m}^{t,e,*}\nabla f_{k,m}(\theta_{k,m}^{t,e-1};\xi_{k,m}^{t,e-1})\right)\right\Vert^2\allowdisplaybreaks\nonumber\\
        =&\left(1+\frac{1}{2E-1}\right)\left\Vert\theta_{k,m}^{t,e-1} - \theta_{k}^t\right\Vert^2 + 4E\eta_L^2\left\Vert\sum_{i=1}^kw_{i,m}^{t,e-1,*}\nabla f_i(\theta_K^t)\right\Vert^2\allowdisplaybreaks\nonumber\\
        &+ 4E\eta_L^2\left\Vert\sum_{i=1}^{k-1}w_{k,m}^{t,e-1,*} \tilde{\boldsymbol{H}}_{i}\left(\theta_{k,m}^{t,e-1} - \theta_{k}^t\right)\right\Vert^2\allowdisplaybreaks\nonumber\\
        \leq&\left(1+\frac{1}{2E-1}\right)\left\Vert\theta_{k,m}^{t,e-1} - \theta_{k}^t\right\Vert^2 + 4\eta_L^2ED^2 + 4\eta_L^2E\left(k-1\right)^2L^2\left\Vert\theta_{k,m}^{t,e-1} - \theta_{k}^t\right\Vert^2\allowdisplaybreaks\nonumber\\
        =&\left(1+\frac{1}{2E-1} + 4\eta_L^2E\left(k-1\right)^2L^2\right)\left\Vert\theta_{k,m}^{t,e-1} - \theta_{k}^t\right\Vert^2 + 4\eta_L^2ED^2.
    \end{align}
    When $\eta_L\leq\frac{1}{3(k-1)EL}$, we have $1+\frac{1}{2E-1} + 4\eta_L^2E\left(k-1\right)^2L^2\leq 1+\frac{1}{E-1}$. Then, we have 
    \begin{equation}
        \left\Vert\theta_{k,m}^{t,e} - \theta_k^t \right\Vert^2\leq\left(1+\frac{1}{E-1} \right)\left\Vert\theta_{k,m}^{t,e-1} - \theta_{k}^t\right\Vert^2 + 4\eta_L^2EK^2D^2.
    \end{equation}
    Unrolling the recursion, we have 
    \begin{equation}
        \left\Vert\theta_{k,m}^{t,e} - \theta_k^t \right\Vert^2\leq\sum_{\tau =0}^{e-1}\left(1+\frac{1}{E-1}\right)^\tau\left(4\eta_L^2ED^2\right)\leq 20\eta_L^2E^2D^2.
    \end{equation}
\end{proof}

\begin{lemma}\label{lemma:partical participants}
Assume we randomly select $N$ clients from all $M$ clients, and the set $\mathcal{S}$ is the set of the client index. If $\varepsilon_n\in\mathbb{R}^d, n\in[M]$ satisfies $\mathbb{E}\left\Vert\varepsilon_n\right\Vert^2\leq \sigma^2$ and $\sum_{n=1}^M\varepsilon_n = 0$, then 
\begin{itemize}
    \item if $\mathbb{E}\left[\varepsilon_n\right] = 0$, $\forall n\in[M]$, we have:
    \begin{equation}
        \mathbb{E}\left\Vert\frac{1}{N}\sum_{n\in\mathcal{S}} \varepsilon_n\right\Vert^2 \leq \frac{\sigma^2}{N}. 
    \end{equation}
    \item if $\mathbb{E}\left[\varepsilon_n\right] \neq 0$, $\forall n\in[M]$, we have:
    \begin{equation}
        \mathbb{E}\left\Vert\frac{1}{N}\sum_{n\in\mathcal{S}} \varepsilon_n\right\Vert^2 \leq \frac{M-N}{N\left(M-1\right)}\sigma^2. 
    \end{equation}
\end{itemize}
\end{lemma}

\begin{proof}
    Since $\sum_{n\in\mathcal{S}}\varepsilon_n = \sum_{n=1}^M \mathbb{I}\{n\in\mathcal{S}\} \varepsilon_n$, we can expand as:
    \begin{align}
        \mathbb{E}\left\Vert\frac{1}{N}\sum_{n\in\mathcal{S}} \varepsilon_n\right\Vert^2 =& \mathbb{E}\left\Vert\frac{1}{N}\sum_{n=1}^M \mathbb{I}\{n\in\mathcal{S}\} \varepsilon_n\right\Vert^2\allowdisplaybreaks\nonumber\\
        =& \frac{1}{N^2}\sum_{n=1}^M\mathbb{E}\left[\mathbb{I}\{n\in\mathcal{S}\}\right] \mathbb{E}\left\Vert\varepsilon_n\right\Vert^2 + \frac{1}{N^2}\sum_{m\neq n}\mathbb{E}\left[\mathbb{I}\{m,n\in\mathcal{S}\}\right]\mathbb{E}\left\langle\varepsilon_n, \varepsilon_m\right\rangle\allowdisplaybreaks\nonumber\\
        =& \frac{1}{MN}\sum_{n=1}^M \mathbb{E}\left\Vert\varepsilon_n\right\Vert^2 + \frac{N-1}{MN\left(M-1\right)}\sum_{m\neq n}\mathbb{E}\left\langle\varepsilon_n, \varepsilon_m\right\rangle
    \end{align}

    Since $\sum_{n=1}^M\varepsilon_n = 0$, then we have 
    \begin{equation}
        \left\Vert\sum_{n=1}^M\varepsilon_n\right\Vert^2 = \left\langle\sum_{n=1}^M\varepsilon_n, \sum_{m=1}^M\varepsilon_m\right\rangle = \sum_{n=1}^M\left\Vert\varepsilon_n\right\Vert^2 + \sum_{m\neq n}\left\langle\varepsilon_n, \varepsilon_m\right\rangle = 0,
    \end{equation}
    where we can get the relation that $\sum_{m\neq n}\left\langle\varepsilon_n, \varepsilon_m\right\rangle = -\sum_{n=1}^M\left\Vert\varepsilon_n\right\Vert^2$. Then, we have:
    \begin{align}
        \mathbb{E}\left\Vert\frac{1}{N}\sum_{n\in\mathcal{S}} \varepsilon_n\right\Vert^2 =&  \left(\frac{1}{MN} - \frac{N-1}{MN\left(M-1\right)}\right)\sum_{n=1}^M \mathbb{E}\left\Vert\varepsilon_n\right\Vert^2\\
        =&\frac{M-N}{MN\left(M-1\right)}\sum_{n=1}^M\mathbb{E}\left\Vert\varepsilon_n\right\Vert^2\\
        \leq& \frac{M-N}{N\left(M-1\right)}\sigma^2.
    \end{align}
    If $\mathbb{E}\left[\varepsilon_n\right] = 0$, $\forall n\in[M]$, we have:
    \begin{align}
        \mathbb{E}\left\Vert\frac{1}{N}\sum_{n\in\mathcal{S}} \varepsilon_n\right\Vert^2 =& 
        \frac{1}{MN}\sum_{n=1}^M \mathbb{E}\left\Vert\varepsilon_n\right\Vert^2\leq \frac{1}{N}\sigma^2.
    \end{align}
    This completes the proof.
\end{proof}

\begin{lemma}\label{lemma:difference of d}
    By denoting $\bar{\boldsymbol{\mathrm{d}}}_{K,m}^{t,e} = \mathbb{E}_t[\tilde{\boldsymbol{\mathrm{d}}}_{K,m}^{t,e}]$, we can bound the norm of the difference between $\tilde{\boldsymbol{\mathrm{d}}}_{K,m}^{t,e}$ and $\mathbb{E}_t[{\boldsymbol{\mathrm{d}}}_{K,m}^{t,e}]$ as
    \begin{equation}
        \mathbb{E}_t\left\Vert \tilde{\boldsymbol{\mathrm{d}}}_{K,m}^{t,e} - \bar{\boldsymbol{\mathrm{d}}}_{K,m}^{t,e}\right\Vert^2\leq \sigma_L^2 + \frac{M-N}{N(M-1)}\beta\sigma_G^2.
    \end{equation}
\end{lemma}

\begin{proof}
\begin{align}
    &\mathbb{E}_t\left\Vert \tilde{\boldsymbol{\mathrm{d}}}_{K,m}^{t,e} - \bar{\boldsymbol{\mathrm{d}}}_{K,m}^{t,e}\right\Vert^2\allowdisplaybreaks\nonumber\\
    =& \mathbb{E}_t\left\Vert \sum_{i=1}^{K-1}w_{i,m}^{t,e,*}\left(\frac{1}{N}\sum_{n\in\mathcal{S}_K^t} \nabla f_{i,n}\left(\theta_K^t;\xi_{i,n}^t\right) + \tilde{\boldsymbol{H}}_i^t\left(\theta_{K,m}^{t,e} - \theta_{K,m}^t\right)\right) + w_{K,m}^{t,e,*}\nabla f_{K,m}\left(\theta_{K,m}^{t,e};\xi_{K,m}^{t,e}\right) \right.\allowdisplaybreaks\nonumber\\
    &\left.- \sum_{i=1}^{K-1}w_{i,m}^{t,e,*}\left(\nabla f_{i}\left(\theta_K^t\right) + \tilde{\boldsymbol{H}}_i^t\left(\theta_{K,m}^{t,e} - \theta_{K,m}^t\right) \right) - w_{K,m}^{t,e,*}\nabla f_{K,m}\left(\theta_{K,m}^{t,e}\right)\right\Vert^2\allowdisplaybreaks\nonumber\\
    =&\mathbb{E}_t\left\Vert\sum_{i=1}^{K-1}w_{i,m}^{t,e,*} \left(\frac{1}{N}\sum_{n\in\mathcal{S}_K^t}\left(\nabla f_{i,n}\left(\theta_{K}^t;\xi_{i,n}^t\right) -\nabla f_{i,n}\left(\theta_K^t\right) + \nabla f_{i,n}\left(\theta_K^t\right) - \nabla f_i\left(\theta_K^t\right)\right)\right)\right.\allowdisplaybreaks\nonumber\\
    &\left.+ w_{K,m}^{t,e,*}\left(\nabla f_{K,m}\left(\theta_{K,m}^{t,e};\xi_{K,m}^{t,e}\right) - \nabla f_{K,m}\left(\theta_{K,m}^{t,e}\right)\right)  \right\Vert^2.
\end{align}
Since 
\begin{align}
        &\mathbb{E}_t\left\Vert \frac{1}{N}\sum_{n\in\mathcal{S}_K^t}\left(\nabla f_{i,n}\left(\theta_{K}^t;\xi_{i,n}^t\right) -\nabla f_{i,n}\left(\theta_K^t\right) + \nabla f_{i,n}\left(\theta_K^t\right) - \nabla f_i\left(\theta_K^t\right)\right)\right\Vert^2 \allowdisplaybreaks\nonumber\\
        \leq & \mathbb{E}_t\left\Vert \frac{1}{N}\sum_{n\in\mathcal{S}_K^t}\left(\nabla f_{i,n}\left(\theta_{K}^t;\xi_{i,n}^t\right) -\nabla f_{i,n}\left(\theta_K^t\right)\right)\right\Vert^2\allowdisplaybreaks\nonumber\\
        &+ \mathbb{E}_t\left\Vert \frac{1}{N}\sum_{n\in\mathcal{S}_K^t}\left(\nabla f_{i,n}\left(\theta_K^t\right) - \nabla f_i\left(\theta_K^t\right)\right)\right\Vert^2\allowdisplaybreaks\nonumber\\
        \leq & \frac{\sigma_L^2}{N} + \frac{M - N}{N\left(M-1\right)}{ \beta}\sigma_G^2,
\end{align}
where the second inequality holds due to Lemma \ref{lemma:partical participants} based on the truth that 
\begin{itemize}
    \item {we have $\mathbb{E}_t\left[\nabla f_{i,n}\left(\theta_{K}^t;\xi_{i,n}^t\right) -\nabla f_{i,n}\left(\theta_K^t\right)\right] =0, \mathbb{E}_t\Vert\nabla f_{i,n}(\theta_{K}^t;\xi_{i,n}^t) - \nabla f_{i,n}(\theta_{K}^t)\Vert^2\leq \sigma_L^2,\forall n\in\mathcal{S}_K^t$ based on Assumption~\ref{assumption4:Bounded Local Gradient Variance}}.
    \item {we have $\mathbb{E}_t\left[\nabla f_{i,n}\left(\theta_{K}^t\right) -\nabla f_{i}\left(\theta_K^t\right)\right]\neq 0,\mathbb{E}_t\Vert\nabla f_{i,n}\left(\theta_{K}^t\right) -\nabla f_{i}\left(\theta_K^t\right)\Vert^2\leq \beta\sigma_G^2, \forall n\in\mathcal{S}_K^t$ based on Assumption~\ref{assumption5:Bounded Global Gradient Variance}}.
\end{itemize}

Therefore, we have 
\begin{align}
    \mathbb{E}_t\left\Vert \tilde{\boldsymbol{\mathrm{d}}}_{K,m}^{t,e} - \bar{\boldsymbol{\mathrm{d}}}_{K,m}^{t,e}\right\Vert^2\leq \sigma_L^2 + \frac{M-N}{N(M-1)}\beta\sigma_G^2.
\end{align}
\end{proof}

\section{Complete Proofs}\label{sec:complete proofs}

\subsection{Proof of Theorem \ref{theorem:BKT}}
\begin{proof}
    We can expand $\mathbb{E}_t\left[ f_{1:K-1}\left(\theta_K^t\right)\right]$ as 
    \begin{align}
            &\mathbb{E}_t\left[ f_{1:K-1}\left(\theta_K^t\right)\right]\allowdisplaybreaks\nonumber\\
            \leq& f_{1:K-1}\left(\theta_K^0\right) +\left\langle \nabla f_{1:K-1}\left(\theta_K^0\right),\mathbb{E}_t\left[\theta_K^t - \theta_K^0\right]\right\rangle + \frac{\left(K-1\right)L}{2}\mathbb{E}_t\left\Vert\theta_K^t - \theta_K^0\right\Vert^2\allowdisplaybreaks\nonumber\\
            =& f_{1:K-1}\left(\theta_K^0\right) - \eta_G\eta_L\left\langle \nabla f_{1:K-1}\left(\theta_K^0\right), \mathbb{E}_t\left[\frac{1}{N}\sum_{\tau=0}^{t-1}\sum_{m\in\mathcal{S}_K^\tau}\sum_{e=0}^{E-1}\tilde{\boldsymbol{\mathrm{d}}}_{K,m}^{\tau,e}\right]\right\rangle\allowdisplaybreaks\nonumber\\
            &+\frac{\left(K-1\right)L}{2}\eta_G^2\eta_L^2\mathbb{E}_t\left\Vert\frac{1}{N}\sum_{\tau=0}^{t-1}\sum_{m\in\mathcal{S}_K^\tau}\sum_{e=0}^{E-1}\tilde{\boldsymbol{\mathrm{d}}}_{K,m}^{\tau,e}\right\Vert^2\allowdisplaybreaks\nonumber\\
            =&f_{1:K-1}\left(\theta_K^0\right) \underbrace{- \eta_G\eta_L\left\langle \nabla f_{1:K-1}\left(\theta_K^0\right), \frac{1}{N}\sum_{\tau=0}^{t-1}\sum_{m\in\mathcal{S}_K^\tau}\sum_{e=0}^{E-1}\bar{\boldsymbol{\mathrm{d}}}_{K,m}^{\tau,e}\right\rangle}_{T_1}\allowdisplaybreaks\nonumber\\
            &+\underbrace{\frac{\left(K-1\right)L}{2}\eta_G^2\eta_L^2\mathbb{E}_t\left\Vert\frac{1}{N}\sum_{\tau=0}^{t-1}\sum_{m\in\mathcal{S}_K^\tau}\sum_{e=0}^{E-1}\tilde{\boldsymbol{\mathrm{d}}}_{K,m}^{\tau,e}\right\Vert^2}_{T_2}.
    \end{align}

Term $T_1$ can be bounded by 
\begin{align}
    T_1 =& - \eta_G\eta_L\left\langle \nabla f_{1:K-1}\left(\theta_K^0\right), \frac{1}{M}\sum_{\tau=0}^{t-1}\sum_{m\in\mathcal{S}_K^\tau}\sum_{e=0}^{E-1}\bar{\boldsymbol{\mathrm{d}}}_{K,m}^{\tau,e}\right\rangle\allowdisplaybreaks\nonumber\\
    \leq& -\frac{\eta_G\eta_L}{M}\cdot\epsilon\left\Vert\nabla f_{1:K-1}\left(\theta_K^0\right)\right\Vert\cdot\left[ \sum_{\tau=0}^{t-1}\sum_{m\in\mathcal{S}_K^\tau}\sum_{e=0}^{E-1}\left\Vert\bar{\boldsymbol{\mathrm{d}}}_{K,m}^{\tau,e}\right\Vert\right].
\end{align}

Term $T_2$ can be bounded as
\begin{align}
    T_2\leq& \frac{\left(K-1\right)L}{2}\cdot\frac{\eta_G^2\eta_L^2Et}{N}\left[\sum_{\tau=0}^{t-1}\sum_{m\in\mathcal{S}_K^\tau}\sum_{e=0}^{E-1}\mathbb{E}_t\left\Vert\tilde{\boldsymbol{\mathrm{d}}}_{K,m}^{\tau,e}\right\Vert^2\right]\allowdisplaybreaks\nonumber\\
    \leq & \underbrace{\frac{\left(K-1\right)L}{2}\cdot\frac{\eta_G^2\eta_L^2Et}{N}\left[\sum_{\tau=0}^{t-1}\sum_{m\in\mathcal{S}_K^\tau}\sum_{e=0}^{E-1}\mathbb{E}_t\left\Vert\bar{\boldsymbol{\mathrm{d}}}_{K,m}^{\tau,e}\right\Vert^2\right]}_{T_{21}}\allowdisplaybreaks\nonumber\\
    +&\underbrace{\frac{\left(K-1\right)L}{2}\cdot\frac{\eta_G^2\eta_L^2Et}{N}\left[\sum_{\tau=0}^{t-1}\sum_{m\in\mathcal{S}_K^\tau}\sum_{e=0}^{E-1}\mathbb{E}_t\left\Vert\bar{\boldsymbol{\mathrm{d}}}_{K,m}^{\tau,e} - \tilde{\boldsymbol{\mathrm{d}}}_{K,m}^{\tau,e}\right\Vert^2\right]}_{T_{22}}.
\end{align}

Since $\eta_L\leq \frac{2\epsilon\left\Vert\nabla f_{1:K-1}\left(\theta_K^0\right)\right\Vert}{DLEt\sqrt{7tEN}}$ and $\eta_G\leq \frac{1}{K-1}$, we have 
\begin{align}
    T_{21} =& \frac{\eta_G\eta_L}{N}\cdot\left(K-1\right)\eta_G\cdot\frac{\eta_LLEt}{2}\left[\sum_{\tau=0}^{t-1}\sum_{m\in\mathcal{S}_K^\tau}\sum_{e=0}^{E-1}\mathbb{E}_t\left\Vert\bar{\boldsymbol{\mathrm{d}}}_{K,m}^{\tau,e}\right\Vert^2\right]\allowdisplaybreaks\nonumber\\
    \leq&\frac{\eta_G\eta_L}{N}\cdot\frac{\epsilon\left\Vert\nabla f_{1:K-1}\left(\theta_K^0\right)\right\Vert}{D\sqrt{7tEN}}\left[\sum_{\tau=0}^{t-1}\sum_{m\in\mathcal{S}_K^\tau}\sum_{e=0}^{E-1}\mathbb{E}_t\left\Vert\bar{\boldsymbol{\mathrm{d}}}_{K,m}^{\tau,e}\right\Vert^2\right]\allowdisplaybreaks\nonumber\\
    \leq& \frac{\eta_G\eta_L}{N}\cdot\epsilon\left\Vert\nabla f_{1:K-1}\left(\theta_K^0\right)\right\Vert\cdot\frac{\sum_{\tau=0}^{t-1}\sum_{m\in\mathcal{S}_K^\tau}\sum_{e=0}^{E-1}\mathbb{E}_t\left\Vert\bar{\boldsymbol{\mathrm{d}}}_{K,m}^{\tau,e}\right\Vert^2}{\sqrt{\sum_{\tau=0}^{t-1}\sum_{m\in\mathcal{S}_K^\tau}\sum_{e=0}^{E-1}\mathbb{E}_t\left\Vert\bar{\boldsymbol{\mathrm{d}}}_{K,m}^{\tau,e}\right\Vert^2}}\allowdisplaybreaks\nonumber\\
    \leq &\frac{\eta_G\eta_L}{N}\cdot\epsilon\left\Vert\nabla f_{1:K-1}\left(\theta_K^0\right)\right\Vert\cdot\sqrt{\sum_{\tau=0}^{t-1}\sum_{m\in\mathcal{S}_K^\tau}\sum_{e=0}^{E-1}\mathbb{E}_t\left\Vert\bar{\boldsymbol{\mathrm{d}}}_{K,m}^{\tau,e}\right\Vert^2}\allowdisplaybreaks\nonumber\\
    \leq &\frac{\eta_G\eta_L}{N}\cdot\epsilon\left\Vert\nabla f_{1:K-1}\left(\theta_K^0\right)\right\Vert\cdot\left[\sum_{\tau=0}^{t-1}\sum_{m\in\mathcal{S}_K^\tau}\sum_{e=0}^{E-1}\mathbb{E}_t\left\Vert\bar{\boldsymbol{\mathrm{d}}}_{K,m}^{\tau,e}\right\Vert\right],
\end{align}
where the second inequality holds due to the truth that 
\begin{align}
    &\mathbb{E}_t\left\Vert\bar{\boldsymbol{\mathrm{d}}}_{K,m}^{\tau,e}\right\Vert^2 \allowdisplaybreaks\nonumber\\
    =& \mathbb{E}_t\left\Vert\sum_{i=1}^{K-1} w_{i,m}^{t,e,*}\left(\nabla f_i\left(\theta_K^t\right) +  \tilde{\boldsymbol{H}}_i^t\left(\theta_{K,m}^{\tau,e} - \theta_{K,m}^t\right)\right) + w_{K,m}^{t,e,*}\nabla f_{K,m}\left(\theta_{K,m}^{\tau,e}\right)\right\Vert^2\allowdisplaybreaks\nonumber\\
    \leq & 2\mathbb{E}_t\left\Vert\sum_{i=1}^{K-1} w_{i,m}^{t,e,*}\nabla f_i\left(\theta_K^t\right)  + w_{K,m}^{t,e,*}\nabla f_{K,m}\left(\theta_{K,m}^{\tau,e}\right)\right\Vert^2\allowdisplaybreaks\nonumber\\& + 2\mathbb{E}_t\left\Vert\sum_{i=1}^{K-1} w_{i,m}^{t,e,*} \tilde{\boldsymbol{H}}_i^\tau\left(\theta_{K,m}^{\tau,e} - \theta_{K,m}^\tau\right)\right\Vert^2\allowdisplaybreaks\nonumber\\
    \leq& 2D^2 + 2L^2\cdot 20\eta_L^2E^2D^2\allowdisplaybreaks\nonumber\\
    \leq& 2D^2+\frac{40D^2}{9\left(K-1\right)^2}\allowdisplaybreaks\nonumber\\
    \leq& 7D^2,
\end{align}
where the second inequality comes from Assumption \ref{assumption1:Bounded Gradient} and Lemma \ref{lemma:bounded local updates}, and the third inequality holds when $\eta_L\leq\frac{1}{3\left(K-1\right)EL}$.

According to Lemma \ref{lemma:difference of d}, term $T_{22}$ can be bounded as 
\begin{align}
    T_{22} =& \frac{\left(K-1\right)L}{2}\cdot\frac{\eta_G^2\eta_L^2Et}{N}\left[\sum_{\tau=0}^{t-1}\sum_{m\in\mathcal{S}_K^\tau}\sum_{e=0}^{E-1}\mathbb{E}_t\left\Vert\bar{\boldsymbol{\mathrm{d}}}_{K,m}^{\tau,e} - \tilde{\boldsymbol{\mathrm{d}}}_{K,m}^{\tau,e}\right\Vert^2\right]\allowdisplaybreaks\nonumber\\
    \leq &\frac{2\epsilon^2\left\Vert\nabla f_{1:K-1}\left(\theta_K^0\right)\right\Vert^2}{7\left(K-1\right)Et^2NLD^2}\left({\sigma_L^2} + \frac{\left(M-N\right){ \beta}\sigma_G^2}{N\left(M-1\right)}\right).
\end{align}

Therefore, we have 
\begin{equation}
        \begin{split}
            \mathbb{E}_t&\left[ f_{1:K-1}\left(\theta_K^t\right)\right]\leq f_{1:K-1}\left(\theta_K^0\right)  + \frac{2\epsilon^2\left\Vert\nabla f_{1:K}\left(\theta_K^0\right)\right\Vert^2}{7\left(K-1\right)Et^2NLD^2}\left({\sigma_L^2} + \frac{\left(M-N\right){ \beta}\sigma_G^2}{N\left(M-1\right)}\right).
        \end{split}
    \end{equation}

\end{proof}

\subsection{Proof of Theorem \ref{theorem:convergence}}

\begin{proof}
According to the update rule, we have
\begin{equation}
    \theta_K^{t+1}  = \theta_K^t + \frac{\eta_G}{N}\sum_{m\in\mathcal{S}_K^t} \Delta_{K,m}^t = \theta_K - \frac{\eta_G\eta_L}{N} \sum_{m\in\mathcal{S}_K^t}\sum_{e=0}^{E-1}\boldsymbol{\mathrm{d}}_{K,m}^{t,e},
\end{equation}
where $\boldsymbol{\mathrm{d}}_{K,m}^{t,e} = \sum_{i=1}^{K-1}w_{i,m}^{t,e,*}(\nabla f_i(\theta_K^t) +  \tilde{\boldsymbol{H}}_i^t(\theta_{K,m}^{t,e} - \theta_K^t)) + w_{K,m}^{t,e,*}\nabla f_{K,m}(\theta_{K,m}^{t,e};\xi_{K,m}^{t,e})$.

Due to the smoothness in Assumption \ref{assumption2:L-smooth}, we have 
\begin{align}\label{eq-conv:global expansion}
            \mathbb{E}_t\left[f_{1:K}\left(\theta_K^{t+1}\right)\right]\leq& f_{1:K}\left(\theta_K^{t}\right) + \underbrace{\mathbb{E}_t\left\langle \nabla f_{1:K}\left(\theta_K^{t}\right),- \frac{\eta_G\eta_L}{M}\sum_{m\in\mathcal{S}_K^t}\sum_{e=0}^{E}\boldsymbol{\mathrm{d}}_{K,m}^{t,e}\right\rangle}_{T_3}\allowdisplaybreaks\nonumber\\
            &+ \frac{KL}{2}\underbrace{\mathbb{E}_t \left\Vert- \frac{\eta_G\eta_L}{N}\sum_{m\in\mathcal{S}_K^t}\sum_{e=0}^{E}\boldsymbol{\mathrm{d}}_{K,m}^{t,e}\right\Vert^2}_{T_4}.
    \end{align}

For term $T_3$, we have 
\begin{align}\label{eq-conv:term T_1}
        T_3 =&\mathbb{E}_t\left\langle \nabla f_{1:K}\left(\theta_K^{t}\right),- \frac{\eta_G\eta_L}{N}\sum_{m\in\mathcal{S}_K^t}\sum_{e=0}^{E}\left(\sum_{i=1}^{K-1}w_{i,m}^{t,e,*} \left(\nabla f_{i}(\theta_{K}^t) +  \tilde{\boldsymbol{H}}_i^t\left(\theta_{K,m}^{t,e} - \theta_K^t\right)\right)\right.\right.\allowdisplaybreaks\nonumber\\ &\left.\left.+w_{K,m}^{t,e,*}\nabla f_{K,m}\left(\theta_{K,m}^{t,e};\xi_{K,m}^{t,e}\right)\right)\right\rangle\allowdisplaybreaks\nonumber\\
        =&\mathbb{E}_t\left\langle \nabla f_{1:K}\left(\theta_K^{t}\right),- \frac{\eta_G\eta_L}{N}\sum_{m\in\mathcal{S}_K^t}\sum_{e=0}^{E-1}\left(\sum_{i=1}^{K-1}w_{i,m}^{t,e,*} \left(\nabla f_{i}(\theta_{K}^t) +  \tilde{\boldsymbol{H}}_i^t\left(\theta_{K,m}^{t,e} - \theta_K^t\right)\right)\right.\right.\allowdisplaybreaks\nonumber\\ &\left.\left.+w_{K,m}^{t,e,*}\nabla f_{K,m}\left(\theta_{K,m}^{t,e}\right)\right)\right\rangle.
\end{align}

By defining $\tilde{\boldsymbol{\mathrm{d}}}_{K,m}^{t,e} = \sum_{i=1}^{K-1}w_{i,m}^{t,e,*}(\nabla f_i(\theta_K^t) +  \tilde{\boldsymbol{H}}_i^t(\theta_{K,m}^{t,e} - \theta_K^t)) + w_{K,m}^{t,e,*}\nabla f_{K,m}(\theta_{K,m}^{t,e})$, we have

\begin{align}\label{eq-conv:term T_1 expansion}
        T_3 = &\mathbb{E}_t\left\langle \nabla f_{1:K}\left(\theta_K^{t}\right),- \frac{\eta_G\eta_L}{N}\sum_{m\in\mathcal{S}_K^t}\sum_{e=0}^{E-1}\tilde{\boldsymbol{\mathrm{d}}}_{K,m}^{t,e}\right\rangle\allowdisplaybreaks\nonumber\\
        =&\frac{\eta_G\eta_L}{N}\sum_{m\in\mathcal{S}_K^t}\sum_{e=0}^{E-1}\mathbb{E}_t\left\langle \nabla f_{1:K}\left(\theta_K^{t}\right),- \tilde{\boldsymbol{\mathrm{d}}}_{K,m}^{t,e}\right\rangle\allowdisplaybreaks\nonumber\\
        =&\underbrace{\frac{\eta_G\eta_L}{N}\sum_{m\in\mathcal{S}_K^t}\sum_{e=0}^{E-1}\mathbb{E}_t\left\langle \nabla f_{1:K}\left(\theta_K^{t}\right) - \sum_{i=1}^{K-1}\nabla f_i\left(\theta_K^t\right) - \nabla f_{K,m}\left(\theta_{K,m}^{t,e}\right),- \tilde{\boldsymbol{\mathrm{d}}}_{K,m}^{t,e}\right\rangle}_{T_{31}} \allowdisplaybreaks\nonumber\\
        &\underbrace{- \frac{\eta_G\eta_L}{N}\sum_{m\in\mathcal{S}_K^t}\sum_{e=0}^{E-1}\mathbb{E}_t\left\langle \sum_{i=1}^{K-1}\nabla f_i\left(\theta_K^t\right) + \nabla f_{K,m}\left(\theta_{K,m}^{t,e}\right), \tilde{\boldsymbol{\mathrm{d}}}_{K,m}^{t,e}\right\rangle}_{T_{32}}.
\end{align}

For term $T_{31}$, we have

\begin{align}\label{eq-conv:term T_11}
        T_{31} =& \frac{\eta_G\eta_L}{N}\sum_{m\in\mathcal{S}_K^t}\sum_{e=0}^{E-1}\mathbb{E}_t\left\langle \nabla f_{1:K}\left(\theta_K^{t}\right) - \sum_{i=1}^{K-1}\nabla f_i\left(\theta_K^t\right) - \nabla f_{K,m}\left(\theta_{K,m}^{t,e}\right),- \tilde{\boldsymbol{\mathrm{d}}}_{K,m}^{t,e}\right\rangle\allowdisplaybreaks\nonumber\\
        =& \frac{\eta_G\eta_L}{N}\sum_{m\in\mathcal{S}_K^t}\sum_{e=0}^{E-1}\mathbb{E}_t\left\langle \sum_{i=1}^K\nabla f_{i}\left(\theta_K^{t}\right) - \sum_{i=1}^{K-1}\nabla f_i\left(\theta_K^t\right) - \nabla f_{K,m}\left(\theta_{K,m}^{t,e}\right),- \tilde{\boldsymbol{\mathrm{d}}}_{K,m}^{t,e}\right\rangle\allowdisplaybreaks\nonumber\\
        =&  \frac{\eta_G\eta_L}{N}\sum_{m\in\mathcal{S}_K^t}\sum_{e=0}^{E-1}{\mathbb{E}_t\left\langle\nabla f_{K}\left(\theta_K^t\right) - \nabla f_{K,m}\left(\theta_{K,m}^{t,e}\right),- \tilde{\boldsymbol{\mathrm{d}}}_{K,m}^{t,e}\right\rangle}\allowdisplaybreaks\nonumber\\
        \leq & \frac{\eta_G\eta_L}{2N}\sum_{m\in\mathcal{S}_K^t}\sum_{e=0}^{E-1}{\mathbb{E}_t\left\Vert\nabla f_{K}\left(\theta_K^t\right) - \nabla f_{K,m}\left(\theta_{K,m}^{t,e}\right)\right\Vert^2} + \frac{\eta_G\eta_L}{2N}\sum_{m\in\mathcal{S}_K^t}\sum_{e=0}^{E-1}\mathbb{E}_t\left\Vert\tilde{\boldsymbol{\mathrm{d}}}_{K,m}^{t,e}\right\Vert^2\allowdisplaybreaks\nonumber\\
        \leq & \frac{\eta_G\eta_L}{2N}\sum_{m\in\mathcal{S}_K^t}\sum_{e=0}^{E-1}{\left(L^2\mathbb{E}_t\left\Vert\theta_K^t - \theta_{K,m}^{t,e}\right\Vert^2 + \beta\sigma_G^2\right)} + \frac{\eta_G\eta_L}{2N}\sum_{m\in\mathcal{S}_K^t}\sum_{e=0}^{E-1}\mathbb{E}_t\left\Vert\tilde{\boldsymbol{\mathrm{d}}}_{K,m}^{t,e}\right\Vert^2\allowdisplaybreaks\nonumber\\
        \leq & { 10\eta_G\eta_L^3E^3L^2D^2 + \frac{\eta_G\eta_LE}{2}\beta\sigma_G^2}+ \frac{\eta_G\eta_L}{2N}\sum_{m\in\mathcal{S}_K^t}\sum_{e=0}^{E-1}\mathbb{E}_t\left\Vert\tilde{\boldsymbol{\mathrm{d}}}_{K,m}^{t,e}\right\Vert^2,
\end{align}
where the first inequality comes from the truth that $\langle a, b\rangle\leq\frac{1}{2}\Vert a\Vert^2 + \frac{1}{2}\Vert b\Vert^2$, the second inequality holds due to Assumption \ref{assumption2:L-smooth}, and the third inequality comes from Lemma \ref{lemma:bounded local updates}.

For term $T_{32}$, we have

\begin{align}\label{eq-conv:term T_12}
        T_{32} =&\frac{\eta_G\eta_L}{N}\sum_{m\in\mathcal{S}_K^t}\sum_{e=0}^{E-1}\left(\mathbb{E}_t\left\langle \sum_{i=1}^{K-1} \tilde{\boldsymbol{H}}_i^t\left(\theta_{K,m}^{t,e} - \theta_{K,m}^t\right), \tilde{\boldsymbol{\mathrm{d}}}_{K,m}^{t,e}\right\rangle\right.\allowdisplaybreaks\nonumber\\
        & - \left.\mathbb{E}_t\left\langle \sum_{i=1}^{K-1}\nabla f_i\left(\theta_K^t\right) + \nabla f_{K,m}\left(\theta_{K,m}^{t,e}\right) + \sum_{i=1}^{K-1} \tilde{\boldsymbol{H}}_i^t\left(\theta_{K,m}^{t,e} - \theta_{K,m}^t\right), \tilde{\boldsymbol{\mathrm{d}}}_{K,m}^{t,e}\right\rangle\right)
\end{align}

Since $\tilde{\boldsymbol{\mathrm{d}}}_{K,m}^{t,e}$ is a general solution in the convex hull of the vectors $\boldsymbol{g}_{K,m}^{t,e}:=\{\nabla f_i(\theta_K^t) +  \tilde{\boldsymbol{H}}_i(\theta_{K,m}^{t,e} - \theta_{K,m}^t)\}_{i=1}^{K-1}\cup\{\nabla f_{K,m}(\theta_{K,m}^{t,e})\}$, any vector $\boldsymbol{t}$ in the convex hull satisfies that $\left\langle \boldsymbol{t}, \tilde{\boldsymbol{\mathrm{d}}}_{K,m}^{t,e}\right\rangle\geq \left\Vert \tilde{\boldsymbol{\mathrm{d}}}_{K,m}^{t,e}\right\Vert^2$ \cite{yang2023federated}. Therefore, we further have

\begin{align}\label{eq-conv:term T_12 expansion}
        T_{32} \leq & - \frac{\eta_G\eta_L}{N}\sum_{m\in\mathcal{S}_K^t}\sum_{e=0}^{E-1}\sum_{i=1}^{K}\mathbb{E}_t\left\Vert\tilde{\boldsymbol{\mathrm{d}}}_{K,m}^{t,e}\right\Vert^2\allowdisplaybreaks\nonumber\\
        &+ \frac{\eta_G\eta_L}{N}\sum_{m\in\mathcal{S}_K^t}\sum_{e=0}^{E-1}\sum_{i=1}^{K-1} \left(\frac{1}{2}\left\Vert  \tilde{\boldsymbol{H}}_i^t\right\Vert^2\cdot\left\Vert\theta_{K,m}^{t,e} - \theta_{K,m}^t\right\Vert^2 + \frac{1}{2}\left\Vert\tilde{\boldsymbol{\mathrm{d}}}_{K,m}^{t,e}\right\Vert^2\right)\allowdisplaybreaks\nonumber\\
        \leq & - \frac{\eta_G\eta_L}{N}\sum_{m\in\mathcal{S}_K^t}\sum_{e=0}^{E-1}\sum_{i=1}^{K}\mathbb{E}_t\left\Vert\tilde{\boldsymbol{\mathrm{d}}}_{K,m}^{t,e}\right\Vert^2\allowdisplaybreaks\nonumber\\
        &+ \frac{\eta_G\eta_L}{N}\sum_{m\in\mathcal{S}_K^t}\sum_{e=0}^{E-1} \left(\frac{K-1}{2}L^2\cdot 20E^2\eta_L^2D^2 + \frac{1}{2}\sum_{i=1}^{K-1}\left\Vert\tilde{\boldsymbol{\mathrm{d}}}_{K,m}^{t,e}\right\Vert^2\right)\allowdisplaybreaks\nonumber\\
        =& -\frac{\eta_G\eta_LK}{2N}\sum_{m\in\mathcal{S}_K^t}\sum_{e=0}^{E-1}\mathbb{E}_t\left\Vert\tilde{\boldsymbol{\mathrm{d}}}_{K,m}^{t,e}\right\Vert^2 + 10\left(K-1\right)\eta_G\eta_L^3E^3D^2L^2,
\end{align}
where the second inequality holds due to Lemma \ref{lemma:bounded local updates} and Lemma \ref{lemma:Hessian2}

Plugging Eq.~(\ref{eq-conv:term T_11}) and Eq.~(\ref{eq-conv:term T_12 expansion}) back to Eq.~(\ref{eq-conv:term T_1 expansion}), we have

\begin{align}\label{eq-conv:term T_1 expansion ends}
        T_3 \leq& 10\eta_G\eta_L^3E^3{L^2}D^2+ \frac{\eta_G\eta_L}{2N}\sum_{m\in\mathcal{S}_K^t}\sum_{e=0}^{E-1}\mathbb{E}_t\left\Vert\tilde{\boldsymbol{\mathrm{d}}}_{K,m}^{t,e}\right\Vert^2 \allowdisplaybreaks\nonumber\\
        &-\frac{\eta_G\eta_LK}{2N}\sum_{m\in\mathcal{S}_K^t}\sum_{e=0}^{E-1}\mathbb{E}_t\left\Vert\tilde{\boldsymbol{\mathrm{d}}}_{K,m}^{t,e}\right\Vert^2+ 10\left(K-1\right)\eta_G\eta_L^2E^3D^2L^2 \allowdisplaybreaks\nonumber\\
        =& \left(\frac{1}{2} - \frac{K}{2}\right)\frac{\eta_G\eta_L}{N}\sum_{m\in\mathcal{S}_K^t}\sum_{e=0}^{E-1}\mathbb{E}_t\left\Vert\tilde{\boldsymbol{\mathrm{d}}}_{K,m}^{t,e}\right\Vert^2 + { 10\eta_G{K}\eta_L^3E^3D^2L^2}.
\end{align}

For term $T_4$, we have 
\begin{align}\label{eq-conv:term T_2}
    T_4 =& \mathbb{E}_t \left\Vert- \frac{\eta_G\eta_L}{N}\sum_{m\in\mathcal{S}_K^t}\sum_{e=0}^{E}\boldsymbol{\mathrm{d}}_{K,m}^{t,e}\right\Vert^2\allowdisplaybreaks\nonumber\\
    \leq& \frac{\eta_G^2\eta_L^2E}{M}\sum_{m=1}^M\sum_{e=0}^{E-1}\mathbb{E}_t\left\Vert \boldsymbol{\mathrm{d}}_{K,m}^{t,e}\right\Vert^2\allowdisplaybreaks\nonumber\\
    \leq& \frac{\eta_G^2\eta_L^2E}{M}\sum_{m=1}^M\sum_{e=0}^{E-1}\mathbb{E}_t\left\Vert \boldsymbol{\mathrm{d}}_{K,m}^{t,e} - \tilde{\boldsymbol{\mathrm{d}}}_{K,m}^{t,e} + \tilde{\boldsymbol{\mathrm{d}}}_{K,m}^{t,e}\right\Vert^2\allowdisplaybreaks\nonumber\\
    \leq& \frac{\eta_G^2\eta_L^2E}{M}\sum_{m=1}^M\sum_{e=0}^{E-1}{\mathbb{E}_t\left\Vert \boldsymbol{\mathrm{d}}_{K,m}^{t,e} - \tilde{\boldsymbol{\mathrm{d}}}_{K,m}^{t,e}\right\Vert^2} + \frac{\eta_G^2\eta_L^2E}{M}\sum_{m=1}^M\sum_{e=0}^{E-1}\mathbb{E}_t\left\Vert\tilde{\boldsymbol{\mathrm{d}}}_{K,m}^{t,e}\right\Vert^2\allowdisplaybreaks\nonumber\\
    \leq& \eta_G^2\eta_L^2E^2\left({\sigma_L^2} + \frac{M-N}{N\left(M-1\right)}{ \beta}\sigma_G^2\right) + \frac{\eta_G^2\eta_L^2E}{N}\sum_{m\in\mathcal{S}_K^t}\sum_{e=0}^{E-1}\mathbb{E}_t\left\Vert\tilde{\boldsymbol{\mathrm{d}}}_{K,m}^{t,e}\right\Vert^2,
\end{align}
where the last inequality holds due to Lemma \ref{lemma:difference of d}.

Plugging Eq.~(\ref{eq-conv:term T_1 expansion ends}) and Eq.~(\ref{eq-conv:term T_2}) back to Eq.~(\ref{eq-conv:global expansion}), we have

\begin{align}\label{eq-conv:global expansion expansion}
    \mathbb{E}_t&\left[f_{1:K}\left(\theta_K^{t+1}\right)\right]\leq f_{1:K}\left(\theta_K^{t}\right) -  \left(\frac{K}{2} - \frac{1}{2} - \frac{\eta_G\eta_LKEL}{2}\right)\frac{\eta_G\eta_L}{N}\sum_{m\in\mathcal{S}_K^t}\sum_{e=0}^{E-1}\mathbb{E}_t\left\Vert\tilde{\boldsymbol{\mathrm{d}}}_{K,m}^{t,e}\right\Vert^2 \allowdisplaybreaks\nonumber\\
    & + { 10\eta_G\eta_L^3KE^3D^2L^2} + \frac{\eta_G^2\eta_L^2KE^2L}{2}\left({\sigma_L^2} + \frac{M-N}{N\left(M-1\right)}{ \beta}\sigma_G^2\right).
\end{align}

By setting $\frac{K}{2} - \frac{1}{2} - \frac{\eta_G\eta_LKEL}{2}\geq \frac{K}{4}$, that is, $\eta_G\eta_L\leq \frac{1}{6EL}$, we have 

\begin{align}\label{eq-conv:tilde d}
    \frac{1}{N}\sum_{m\in\mathcal{S}_K^t}\sum_{e=0}^{E-1}\mathbb{E}_t\left\Vert\tilde{\boldsymbol{\mathrm{d}}}_{K,m}^{t,e}\right\Vert^2\leq& \frac{4\left(f_{1:K}\left(\theta_K^t\right) - f_{1:K}\left(\theta_K^{t+1}\right)\right)}{\eta_G\eta_LK} +  { 40\eta_L^2KE^3D^2L^2}\allowdisplaybreaks\nonumber\\
    &+ 2\eta_G\eta_LE^2L\left({\sigma_L^2} + \frac{M-N}{N\left(M-1\right)}{\beta}\sigma_G^2\right)
\end{align}

By defining $\nabla \tilde{f}_{1:K}(\theta_K^t) = \sum_{i=1}^K \frac{1}{N}\sum_{m\in\mathcal{S}_K^t}\frac{1}{E}\sum_{e=0}^{E-1}w_{i,m}^{t,e,*}\nabla f_{i}(\theta_K^t)$, we have 
\begin{align}\label{eq-conv:weighted task uniform convergence times E}
    \mathbb{E}_t&\left\Vert E\nabla \tilde{f}_{1:K}(\theta_K^t)\right\Vert^2 = \mathbb{E}_t\left\Vert \sum_{i=1}^K \frac{1}{N}\sum_{m\in\mathcal{S}_K^t}\sum_{e=0}^{E-1}w_{i,m}^{t,e,*}\nabla f_{i}(\theta_K^t)\right\Vert^2\allowdisplaybreaks\nonumber\\
    \leq & 2\mathbb{E}_t\left\Vert \frac{1}{N}\sum_{m\in\mathcal{S}_K^t}\sum_{e=0}^{E-1}\tilde{\boldsymbol{\mathrm{d}}}_{K,m}^{t,e}\right\Vert^2 + 2\underbrace{\mathbb{E}_t\left\Vert \sum_{i=1}^K \frac{1}{N}\sum_{m\in\mathcal{S}_K^t}\sum_{e=0}^{E-1}w_{i,m}^{t,e,*}\nabla f_{i}(\theta_K^t) - \frac{1}{N}\sum_{m\in\mathcal{S}_K^t}\sum_{e=0}^{E-1}\tilde{\boldsymbol{\mathrm{d}}}_{K,m}^{t,e}\right\Vert^2}_{T_3}.
\end{align}

As for the term $T_3$, we have

\begin{align}\label{eq-conv:term T_3}
    T_3
    =& \mathbb{E}_t\left\Vert \sum_{i=1}^{K-1} \frac{1}{N}\sum_{m\in\mathcal{S}_K^t}\sum_{e=0}^{E-1}w_{i,m}^{t,e,*} \tilde{\boldsymbol{H}}_i^t\left(\theta_K^t - \theta_{K,m}^{t,e}\right) \right.\allowdisplaybreaks\nonumber\\
    &\left.+ \frac{1}{N}\sum_{m\in\mathcal{S}_K^t}\sum_{e=0}^{E-1}w_{K,m}^{t,e,*} \left(\nabla f_K\left(\theta_K^t\right) - \nabla f_{K,m}\left(\theta_{K,m}^{t,e}\right)\right)\right\Vert^2\allowdisplaybreaks\nonumber\\
    \leq & 2\underbrace{\mathbb{E}_t\left\Vert \sum_{i=1}^{K-1} \frac{1}{N}\sum_{m\in\mathcal{S}_K^t}\sum_{e=0}^{E-1}w_{i,m}^{t,e,*} \tilde{\boldsymbol{H}}_i^t\left(\theta_K^t - \theta_{K,m}^{t,e}\right) \right\Vert^2}_{T_{31}}\allowdisplaybreaks\nonumber\\
    &+ 2\underbrace{\mathbb{E}_t\left\Vert \frac{1}{N}\sum_{m\in\mathcal{S}_K^t}\sum_{e=0}^{E-1}w_{K,m}^{t,e,*} \left(\nabla f_K\left(\theta_K^t\right) - \nabla f_{K,m}\left(\theta_{K,m}^{t,e}\right)\right)\right\Vert^2}_{T_{32}}.
\end{align}

According to Lemma \ref{lemma:Hessian2}, we have:
\begin{align}\label{eq-conv:term T_31}
    T_{31} \leq& \mathbb{E}_t\left\Vert \sum_{i=1}^{K-1} \frac{1}{N}\sum_{m\in\mathcal{S}_K^t}\sum_{e=0}^{E-1}w_{i,m}^{t,e,*}\right\Vert^2\cdot L^2\cdot 20\eta_L^2E^2D^2\allowdisplaybreaks\nonumber\\
    =& \mathbb{E}_t\left\Vert \frac{1}{N}\sum_{m\in\mathcal{S}_K^t}\sum_{e=0}^{E-1}\left(1-w_{K,m}^{t,e,*}\right)\right\Vert^2\cdot20\eta_L^2E^2D^2 L^2.
\end{align}

Meanwhile, we can also bound term $T_{32}$ as
\begin{align}\label{eq-conv:term T_32}
    T_{32} =& \mathbb{E}_t\left\Vert \frac{1}{N}\sum_{m\in\mathcal{S}_K^t}\sum_{e=0}^{E-1}w_{K,m}^{t,e,*} \left(\nabla f_{K,m}\left(\theta_K^t\right) - \nabla f_{K,m}\left(\theta_{K,m}^{t,e}\right)\right)\right\Vert^2\allowdisplaybreaks\nonumber\\
    \leq& {\mathbb{E}_t\left\Vert\frac{1}{N}\sum_{m\in\mathcal{S}_K^t}\sum_{e=0}^{E-1}w_{K,m}^{t,e,*}\right\Vert^2 \cdot \left(20\eta_L^2E^2D^2L^2 + \beta\sigma_G^2\right)}\allowdisplaybreaks\nonumber\\
    \leq& {\mathbb{E}_t\left\Vert\frac{1}{N}\sum_{m\in\mathcal{S}_K^t}\sum_{e=0}^{E-1}w_{K,m}^{t,e,*}\right\Vert^2 \cdot 20\eta_L^2E^2D^2L^2 + \beta E\sigma_G^2}
\end{align}

Plugging Eq.~(\ref{eq-conv:term T_31}) and Eq.~(\ref{eq-conv:term T_32}) back to Eq.~(\ref{eq-conv:term T_3}), we have 
{ \begin{align}\label{eq-conv:term T_3 expansion}
     T_3& \leq 40\left(\mathbb{E}_t\left\Vert \frac{1}{N}\sum_{m\in\mathcal{S}_K^t}\sum_{e=0}^{E-1}\left(1-w_{K,m}^{t,e,*}\right)\right\Vert^2 + \mathbb{E}_t\left\Vert\frac{1}{N}\sum_{m\in\mathcal{S}_K^t}\sum_{e=0}^{E-1}w_{K,m}^{t,e,*}\right\Vert^2\right)\eta_L^2E^2D^2L^2\allowdisplaybreaks\nonumber\\
    & + 2\beta E\sigma_G^2.
\end{align}}

Plugging Eq.~(\ref{eq-conv:term T_3 expansion}) back to Eq.~(\ref{eq-conv:weighted task uniform convergence times E}), we have 
\begin{align}
    \mathbb{E}_t&\left\Vert\nabla \tilde{f}_{1:K}\left(\theta_{K}^t\right)\right\Vert^2\leq\frac{2}{E}\cdot\frac{1}{N}\sum_{m\in\mathcal{S}_K^t}\sum_{e=0}^{E-1}\left\Vert\tilde{\boldsymbol{\mathrm{d}}}_{K,m}^{t,e}\right\Vert^2 + 80\left(\mathbb{E}_t\left\Vert \frac{1}{EN}\sum_{m\in\mathcal{S}_K^t}\sum_{e=0}^{E-1}\left(1-w_{K,m}^{t,e,*}\right)\right\Vert^2\right.\allowdisplaybreaks\nonumber\\
    &\quad\quad\quad\quad\quad\quad\left. + \mathbb{E}_t\left\Vert\frac{1}{EN}\sum_{m\in\mathcal{S}_K^t}\sum_{e=0}^{E-1}w_{K,m}^{t,e,*}\right\Vert^2\right)\eta_L^2E^2D^2L^2 { + 4\beta E\sigma_G^2}\allowdisplaybreaks\nonumber\\
    \leq& \frac{8\left(f_{1:K}\left(\theta_K^t\right) - f_{1:K}\left(\theta_K^{t+1}\right)\right)}{\eta_G\eta_LEK} +  80\eta_L^2E^2D^2L^2+ 4\eta_G\eta_LEL\left({\sigma_L^2} + \frac{M-N}{N\left(M-1\right)}{ \beta}\sigma_G^2\right)\allowdisplaybreaks\nonumber\\
    &+ 80\left(\mathbb{E}_t\left\Vert \frac{1}{EN}\sum_{m\in\mathcal{S}_K^t}\sum_{e=0}^{E-1}\left(1-w_{K,m}^{t,e,*}\right)\right\Vert^2{ } + \mathbb{E}_t\left\Vert\frac{1}{EN}\sum_{m\in\mathcal{S}_K^t}\sum_{e=0}^{E-1}w_{K,m}^{t,e,*}\right\Vert^2\right)\eta_L^2E^2D^2L^2 { + 4\beta \sigma_G^2}.
\end{align}

Rearranging and summing from $t = 0,\ldots, T-1$, we have:
\begin{align}
    \frac{1}{T}\sum_{t=0}^{T-1}\mathbb{E}_t\left\Vert\nabla \tilde{f}_{1:K}\left(\theta_{K}^t\right)\right\Vert^2\leq& \frac{8\left(f_{1:K}^0 - f_{1:K}^*\right)}{\eta_G\eta_LEKT} + \frac{4\eta_G\eta_LEL\sigma_L^2}{N} + \frac{4\left(M-N\right)}{N\left(M-1\right)}\eta_G\eta_L{ \beta}EL\sigma_G^2\allowdisplaybreaks\nonumber\\
    & + 80\left[1 + \Lambda_{1:K-1} + {}\Lambda_{1:K} \right]\eta_L^2E^2D^2L^2 { + 4\beta \sigma_G^2},
\end{align}
where $f_{1:K}^0\triangleq f_{1:K}\left(\theta_K^0\right)$, $f_{1:K}^*\triangleq f_{1:K}\left(\theta_K^*\right)$, $\Lambda_K\triangleq \frac{1}{T}\sum_{t=1}^{T}\mathbb{E}_t\left\Vert \frac{1}{EN}\sum_{m\in\mathcal{S}_K^t}\sum_{e=0}^{E-1}w_{K,m}^{t,e,*}\right\Vert^2$, and $\Lambda_{1:K-1} \triangleq \frac{1}{T}\sum_{t=1}^{T}\mathbb{E}_t\left\Vert \frac{1}{EN}\sum_{m\in\mathcal{S}_K^t}\sum_{e=0}^{E-1}\left(1-w_{K,m}^{t,e,*}\right)\right\Vert^2$.
\end{proof}

\section{Details of Experiments}\label{sec:details of experiments}
\subsection{Datasets}
We estimate the performance of \Adaptive and baselines on two class-incremental datasets and two domain-incremental datasets:
\begin{itemize}
    \item Class-incremental datasets:
    \begin{itemize}
        \item \textbf{CIFAR100} consists of $100$ object classes with $50,000$ training images and $10,000$ test images. We randomly divide the $100$ classes into $5$ disjoint tasks, with each task containing $20$ classes.

        \item \textbf{ImageNet100} is a subset of ImageNet containing $100$ object classes, with $50,000$ training images and $5,000$ test images. Similar to CIFAR100, we randomly divide the $100$ classes into $5$ disjoint tasks, with $20$ classes assigned to each task.
    \end{itemize}
    \item Domain-incremental datasets:
    \begin{itemize}
        \item \textbf{Digit-10} consists of $10$ classes across $4$ domains: MNIST, USPS, SVHN, and EMNIST, with a total of $380,548$ images. Similarly, each domain is treated as an individual task, resulting in $4$ sequential tasks.
        
        \item \textbf{VLCS} consists of $5$ classes across $4$ domains: VOC2007, LabelMe, Caltech101, and SUN09, with a total of $7,486$ images. We treat each domain as an individual task, resulting in $4$ sequential tasks.
    \end{itemize}
\end{itemize}

\subsection{Detailed Configurations}\label{sec:detailed configurations}
Our experiments contain two parts: the main experiment based on 4 datasets and the ablation study.
\begin{itemize}
    \item The main experiment focuses on evaluating the performance of \Adaptive in the four datasets and comparing it with other baselines in two metrics, i.e., ACC and BWT. In the main experiment, the local learning rate $\eta_L$ is initialized as $0.001$ and decayed with the rate of $0.96$ after each $5$ epochs, and the global learning rate $\eta_G$ is fixed at $1.0$. We consider the number of clients to be $16$ and the number of participants to be $8$ each round. We set communication rounds $T = 50$ and local { gradient steps} $E = 3$ for all datasets. We set the dirichlet parameter $\alpha = 0.5$ for two domain-incremental datasets and $\alpha = 1.0$ for two class-incremental datasets. The size of the buffer for each client is fixed at $400$. The maximum iterations for Frank Wolfe and Lanczos algorithms are fixed at $10$. { The evaluation on class-incremental benchmarks is based on class-incremental learning paradigm.}
    \item In the ablation study, we conduct a series of experiments to evaluate the performance of \Adaptive under different settings on CIFAR100. In the ablation study of the updating frequency, we set the update interval $\rho = \{1,3,5,10\}$. In the ablation study of buffer size, we set buffers with different sizes $\{100,400,700,1000\}$. In the ablation study of the truncated rank-$r$, we set $r = \{1,3,5,10\}$. In the ablation study on the straggler robustness, we set the ratio $\rho = \{0.25,0.5,0.75,1.0\}$. In the client scaling and participation ratio, we consider two configurations as $N/M = \{8/8,8/16,8/32,8/64\}$ and $N/M = \{2/8,4/16,8/32,16/64\}$. All ablation studies follow the same basic experimental settings as the main experiments, with only the configuration corresponding to the ablation component being varied.
\end{itemize}

 { FedPOCL and FedBECAME are our federated adaptations of centralized continual-learning methods. For the federated adaptation of FedPOCL and FedBECAME, each client's local training follows the original method, while the resulting models are uploaded to the server for aggregation and global evaluation. Upon task completion, FedPOCL updates each client's buffer size using its private dataset, while FedBECAME locally computes $\theta_{k,m}^{GP}$, $\hat{\theta}_{k,m}^{*}$, the precision matrix, and the merged model $\theta_{k,m}^{*}$. These quantities remain private and are not shared with the server. Therefore, to preserve the original mechanisms of both baselines, their federated adaptations primarily introduce server-side aggregation and global evaluation of locally trained models.} In addition, the hyperparameters of baselines are listed in Table~\ref{tab: Hyperparameters for baselines}, and unspecified hyperparameters follow the settings reported in the original papers.

\begin{table*}[h]
\centering
\caption{Hyperparameters for baselines}
    \label{tab: Hyperparameters for baselines}
\begin{tabular}{c|c}
\toprule
Baselines                & Hyperparameters                                                                                          \\ \midrule
FedProx                   & $\lambda = 0.5$                          \\\hline

FLwF-2T                  & $\alpha = 0.5, \beta = 0.25$                                                                             \\\hline
FedSSI & $\lambda = 0.5$\\ \hline
FedAGC & $\gamma = 0.5, \lambda = 0.01$\\ \hline
SPECIAL & $\lambda = 0.4$ for Digit-10 and VLCS, $\lambda = 0.3$ for CIFAR100 and ImageNet100                                                     \\ \bottomrule
\end{tabular}
\end{table*}

{Baseline-specific hyperparameters follow the configurations reported above and the original methods’ recommended settings for unspecified parameters, without an extensive adaptation-specific search. The cross-method results therefore compare these documented configurations under matched training and replay budgets.}

\subsection{Additional Ablation Studies}\label{sec:Additional Ablation Studies}

In order to observe the evolution of model performance on the whole task sequence, we further define the average all-task accuracy for task $i$ by 
\begin{equation}
    \bar{\text{ACC}}_i = \frac{1}{K}\sum_{j=1}^{K}A_{i,j},
\end{equation}
where $A_{i,j}$ represents the test accuracy on task $j$ after learning task $i$. Note that this metric is evaluated over all $K$ tasks, including both seen and unseen tasks at task $i$.

\begin{figure}[htp]
    \centering
    \includegraphics[width=0.8\linewidth]{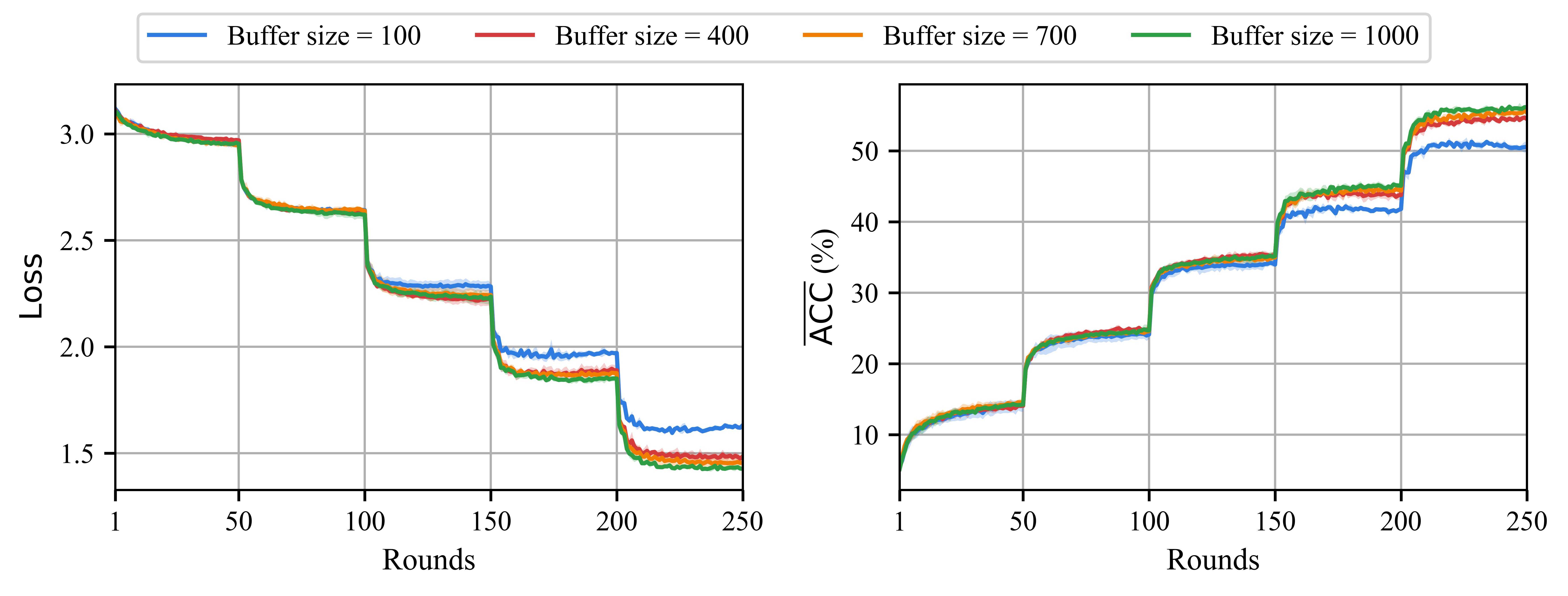} 
    \caption{Loss and ACC ($\%$) of \Adaptive with different buffer sizes}
    \label{fig:buffersize}
\end{figure}

\textbf{Ablation on the buffer size.} Buffer size determines the number of previous samples that a client can store to be used to reconstruct the gradients. Figure~\ref{fig:buffersize} shows that a larger buffer size provides richer previous-information and leads to better performance. However, the performance degradation remains marginal even when the buffer size is reduced to $400$, indicating that \Adaptive does not heavily rely on storing a large number of previous samples and it can still effectively extract and aggregate curvature information even based on a small number of previous samples.

\begin{figure}[h]
    \centering
    \includegraphics[width=0.8\linewidth]{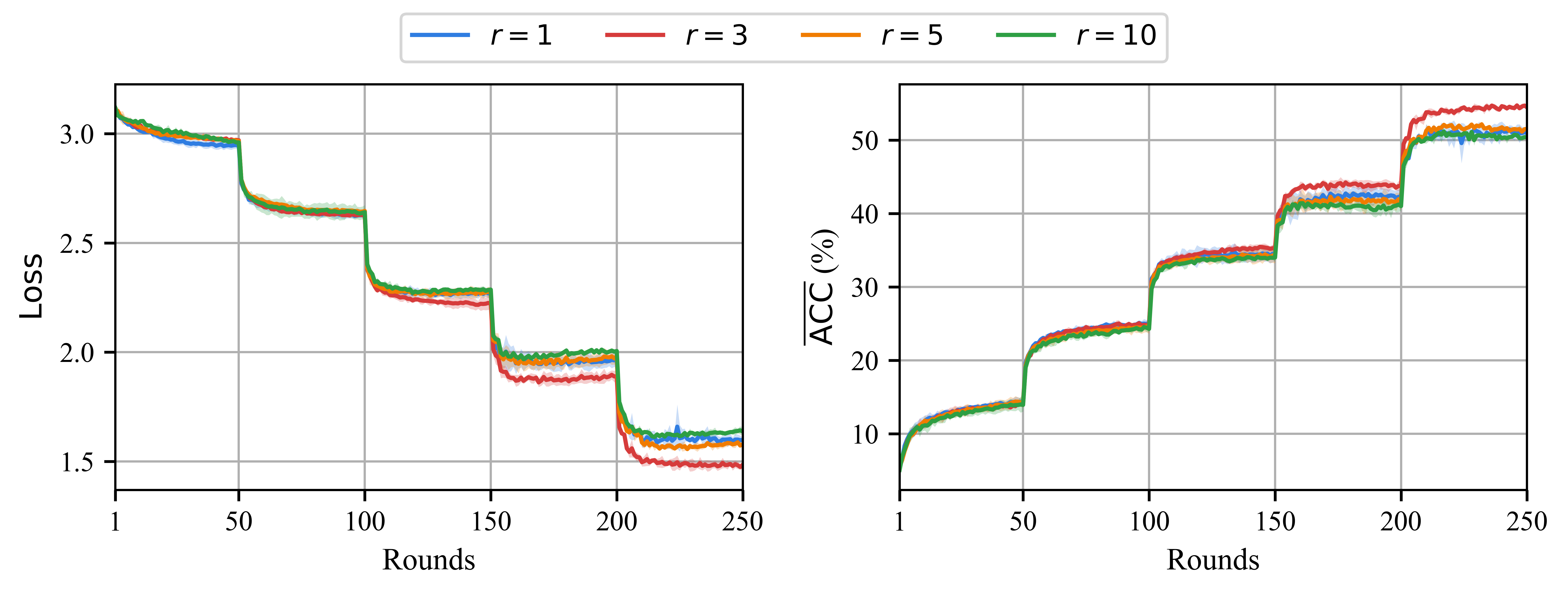} 
    \caption{Loss and ACC ($\%$) of \Adaptive with different truncation ranks $r$}
    \label{fig:rank}
\end{figure}

\textbf{Ablation on the rank-$r$ in truncated EVD.} The value of $r$ determines the number of eigendirections retained for Hessian reconstruction. A larger $r$ preserves curvature information along more directions, but it would  also introduce additional non-dominant information and noise that can impair the model's generalization ability. Figure~\ref{fig:rank} shows that the best performance is achieved when the truncation rank $r$ is set to $3$, indicating that the dominant Hessian information can be effectively captured by only a few leading eigendirections, and excessive eigendirections may introduce additional client-specific variations to degrade the overall performance. 

\begin{figure}[h]
    \centering
    \includegraphics[width=0.8\linewidth]{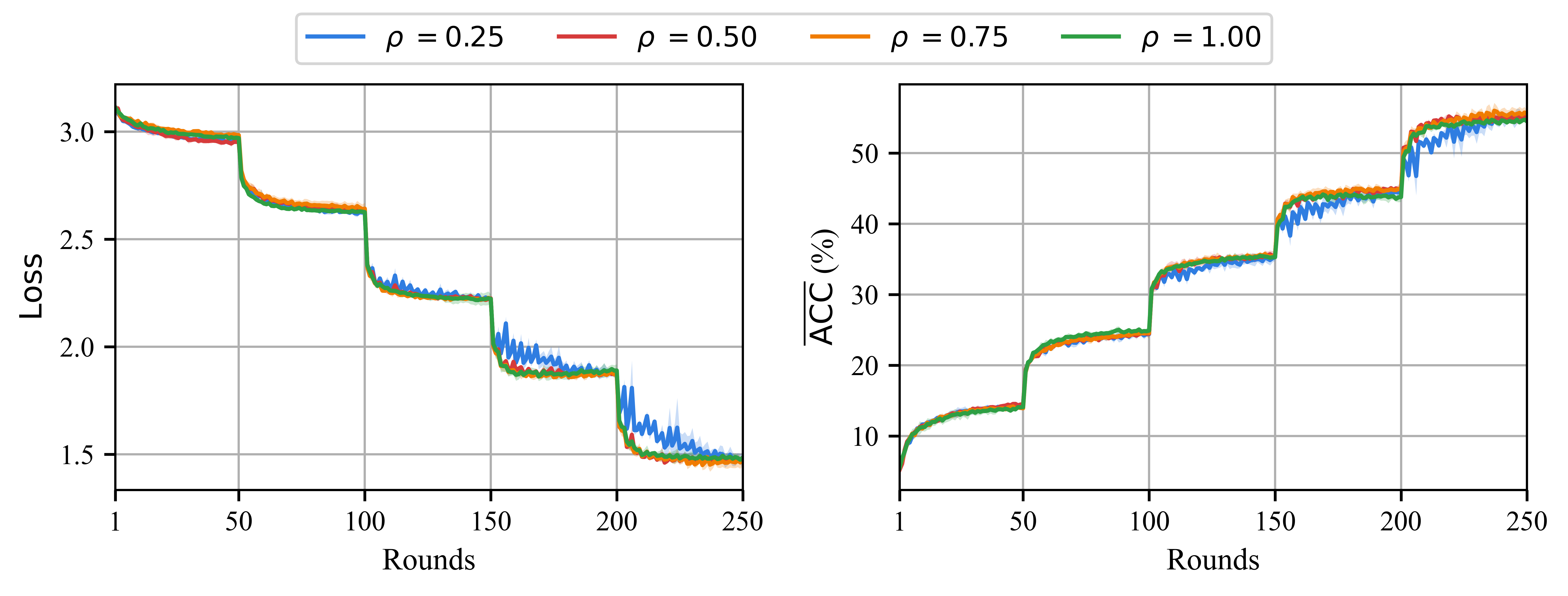} 
    \caption{Loss and ACC ($\%$) of \Adaptive with different fractions of participants uploading the curvature information. }
    \label{fig:partial hessian}
\end{figure}

\textbf{Straggler robustness.} Curvature information exchange requires all participants to upload their previous-task information in our default setting. However, the bandwidth and hardware constraints may prevent stragglers from transmitting their information at every round. Therefore, we consider a restricted setting where only a fraction of participants upload their curvature information to be aggregated in the server. Figure~\ref{fig:partial hessian} shows that the performance degrades only marginally when only $50\%$ of the participants provide curvature information in each round, and the final performance remains competitive even when the fraction is further reduced to $25\%$, demonstrating the robustness of \Adaptive under more practical settings.

\begin{figure}[h]
    \centering
    \includegraphics[width=0.8\linewidth]{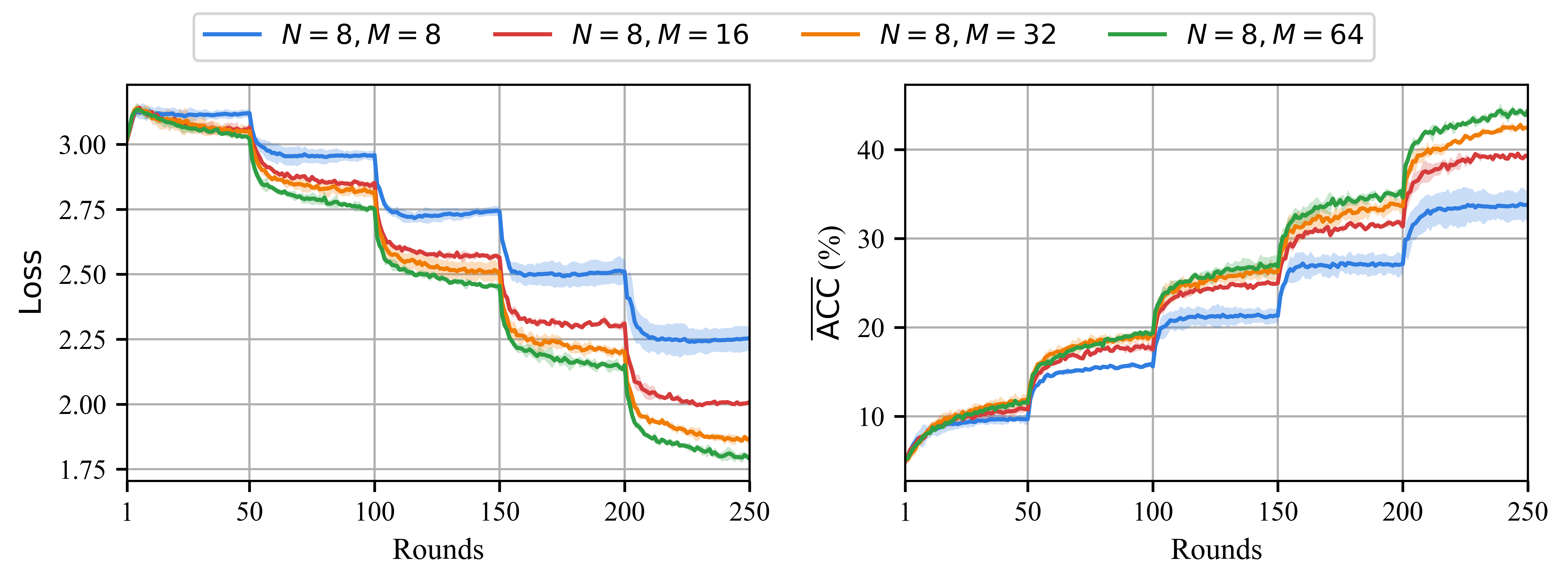} 
    \caption{Loss and ACC ($\%$) of \Adaptive with fixed number of participants but different ratios. }
    \label{fig:participants}
\end{figure}

\begin{figure}[h]
    \centering
    \includegraphics[width=0.8\linewidth]{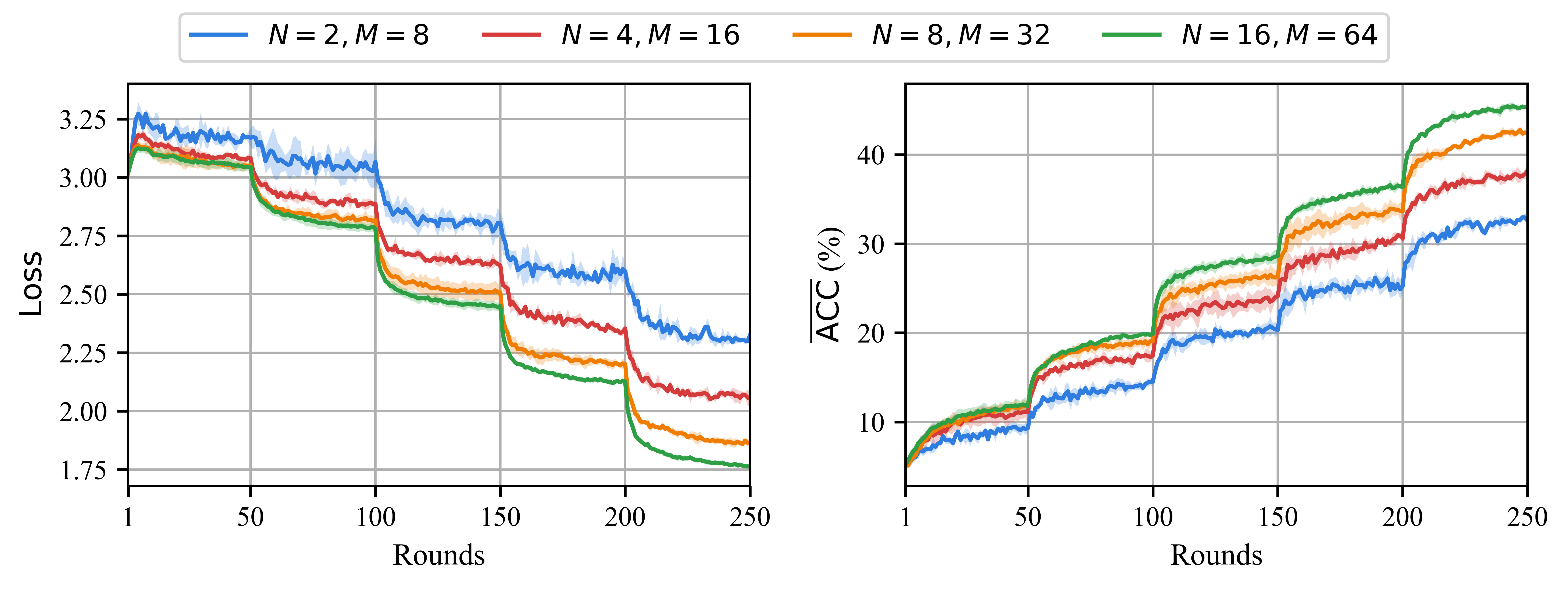} 
    \caption{Loss and ACC ($\%$) of \Adaptive with fixed ratio but different client population sizes. }
    \label{fig:ratio}
\end{figure}

\textbf{Client scaling and participation ratio.} Figure~\ref{fig:participants} and Figure~\ref{fig:ratio} study how \Adaptive scales with the participation ratio and client population when the entire dataset is distributed across $64$ clients. With fixed participants $N=8$, increasing the number of total clients $M$ can expose the model to more client data to improve the performance. With fixed ratio $N/M = 1/4$, scaling up both $M$ and $N$ allows more clients to participate in each training round, enabling the model to benefit from more diverse local information and leading to improved training performance. This observation strengthens~Theorem~\ref{theorem:convergence} and highlights the dominant effect of $M$ on model performance.

\begin{figure}[htp]
    \centering
    \includegraphics[width=0.8\linewidth]{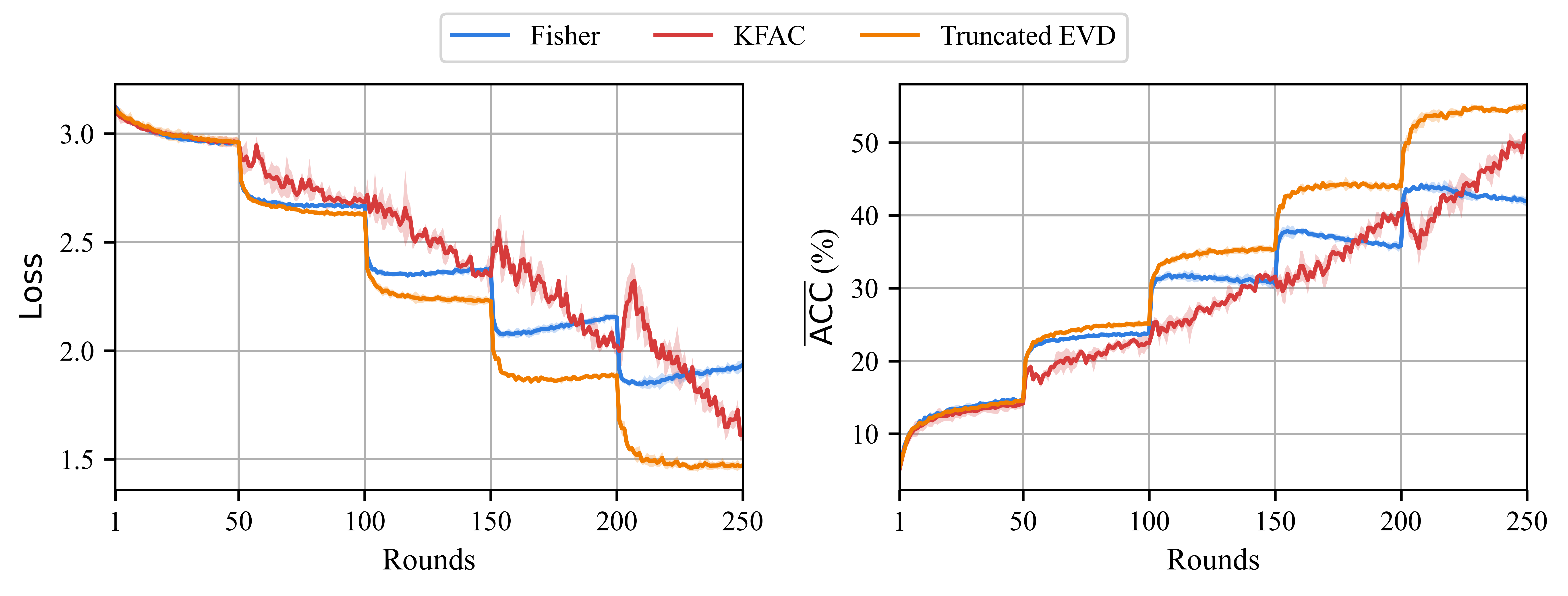} 
    \caption{Different Hessian Approximation.}
    \label{fig:Different Hessian Approximation}
\end{figure}

\textbf{Different Hessian Approximations.} We compare the truncated EVD Hessian approximation method adopted in \Adaptive with the Fisher-information-based method and Kronecker-factored approximate curvature (K-FAC), to evaluate the effectiveness of different Hessian approximation strategies. Figure~\ref{fig:Different Hessian Approximation} shows that the Fisher-information-based approximation can effectively retain previous information during the first few rounds after transitioning to a new task. However, the Fisher approximation may ignore the covariance of gradients which captures the interactions among parameters, limiting the accuracy of curvature approximation and degrading the performance. Although K-FAC captures richer parameter interactions, its approximation accuracy is constrained by the assumed Kronecker and layer-wise structures, which may contribute to insufficient characterization of the dominant curvature directions and affect the stability of model training. In contrast, our truncated EVD mechanism retains the top-$r$ dominant eigendirections, which jointly capture the important covariance and preserve the principal information of the original Hessian, thereby achieving stable performance across all tasks.

{ \textbf{Scalability to Longer Task Sequences.}
To evaluate the scalability of FedGRIT, we extend the CIFAR-100 experiments from 5 tasks to 10 tasks with 10 classes each and 20 tasks with 5 classes each, while maintaining a fixed replay budget of 400 samples per client. We compare FedGRIT with FedPOCL, the strongest baseline in terms of average ACC in the main experiments, under matched training configurations. The detailed results are reported in Table~\ref{tab:longer task sequences}. The results show that FedGRIT consistently outperforms FedPOCL in ACC by $5.75$ and $4.38$ percentage points under $10$ and $20$ tasks, respectively, while reducing total training time by $15.4\%$ and $13.5\%$. FedGRIT also achieves better BWT under $10$ tasks, although FedPOCL exhibits less forgetting under $20$ tasks. As the number of tasks increases, the fixed replay budget must accommodate more historical tasks, introducing greater challenges in knowledge retention. Meanwhile, the communication overhead of FedGRIT increases with the number of tasks due to the exchange of historical gradient and curvature information. Nevertheless, these results demonstrate that FedGRIT maintains its accuracy advantage under longer task sequences, with lower measured training time at the cost of increased communication overhead.

\begin{table*}[h]
    \centering
    \caption{Scalability comparison of FedGRIT and FedPOCL on CIFAR-100 under longer task sequences. The total replay budget is fixed at $400$ samples per client across all configurations. We report final ACC, BWT, total training time, and communication cost.}
    \label{tab:longer task sequences}
    \resizebox{\textwidth}{!}{\begin{tabular}{lccccc}
    \toprule
     Method & dataset split &ACC (\%) $\uparrow$ & BWT (pp) $\uparrow$ &total time (h) $\downarrow$ & Communication Cost (MB) $\downarrow$\\
    \midrule
     FedPOCL & 10*10 &  $30.01$ & $-17.94$ & $14.53$ & $85.73$\\
     FedPOCL & 20*5 &  $22.50$ & $-20.04$ & $28.69$ & $85.73$\\
     \hline
     FedGRIT & 10*10 &  $35.76$ & $-15.37$ & $12.29$ & $1717.95$\\
     FedGRIT & 20*5 &  $26.88$ & $-21.64$ & $24.82$ & $3297.96$\\
    \bottomrule
    \end{tabular}}
\end{table*}}

{
\textbf{Effects of Curvature Correction on Gradient Reconstruction and Learning Performance. } To evaluate the effectiveness of curvature correction, we compare FedGRIT with its curvature-free variant at matched intermediate local models. Specifically, we measure the cosine similarity between reconstructed historical gradients and reference gradients directly evaluated at the same local models, using identical replay samples and aggregation rules. We further evaluate the mean historical-task loss after adaptive weighting, where both variants employ the same weighting mechanism.
\begin{table}[h]
\centering
\caption{Effects of curvature correction on historical-gradient
reconstruction and historical-task loss at different checkpoints.
Higher cosine similarity and lower loss indicate better performance.}
\label{tab:curvature_diagnostics}
\begin{tabular}{cccccc}
\toprule
\multirow{2}{*}{Task} &
\multirow{2}{*}{Round} &
\multicolumn{2}{c}{No Curvature} &
\multicolumn{2}{c}{FedGRIT} \\
\cline{3-6}
& & Similarity (cos) & Loss & Similarity (cos) & Loss \\
\midrule
$4$ & $2$  & $0.215$ & $1.9403$ & $0.244$ & $1.8869$ \\
$4$ & $48$ & $0.559$ & $1.9840$ & $0.663$ & $1.9403$ \\
$5$ & $2$  & $0.247$ & $2.1352$ & $0.328$ & $2.0888$ \\
$5$ & $48$ & $0.543$ & $2.1120$ & $0.706$ & $2.0800$ \\
\bottomrule
\end{tabular}
\end{table}
As shown in Table~\ref{tab:curvature_diagnostics}, curvature correction consistently improves gradient cosine similarity by 0.029--0.163 and reduces historical-task loss by 0.0320--0.0534 across all evaluated checkpoints. These results demonstrate that curvature information improves historical-gradient reconstruction at evolving local models, leading to more effective adaptive updates. Furthermore, the end-to-end curvature ablation yields improvements of 14.62 percentage points in ACC and 7.90 percentage points in BWT, further validating the contribution of curvature correction to overall learning performance.
}

{

\textbf{Detailed Hyperparameter Sensitivity.} We further investigate the effects of replay buffer size, curvature rank, and refresh interval on FedGRIT's performance and computational overhead. The detailed results in Table~\ref{tab:hyperparameter_ablation} show that a refresh interval of 3 achieves the highest ACC ($54.66\%$) and best BWT ($-6.07$ pp), reducing runtime by $29.9\%$ compared with an interval of $1$. Similarly, rank $3$ achieves the best ACC and BWT while avoiding the substantial computational and memory overhead associated with higher ranks. Therefore, we select a refresh interval of 3 and curvature rank of 3 to balance learning performance and computational efficiency. As for the selection of replay buffer size,  by increasing the replay budget from $100$ to $400$ samples, ACC and BWT are increased by $4.07$ pp and $4.06$ pp respectively. Although larger buffers further improve performance, 400 samples achieve ACC within $1.53$ pp of the 1,000-sample setting while retaining $60\%$ fewer samples and reducing per-round runtime by $20.9\%$. Therefore, we adopt a total replay budget of 400 samples per client as a practical trade-off between performance and resource consumption.

\begin{table}[h]
\centering
\caption{Effects of curvature refresh interval, approximation rank,
and replay buffer size on FedGRIT's performance and computational
overhead on CIFAR-100.}
\label{tab:hyperparameter_ablation}
\begin{tabular}{cccccc}
\toprule
Type & Setting & ACC (\%) $\uparrow$ & BWT (pp) $\uparrow$
& Time (s) $\downarrow$ & Peak Memory Usage (GB) $\downarrow$ \\
\midrule
\multirow{4}{*}{Frequency}
& $1$  & $50.66$ & $-10.26$ & $91.38$ & $18.94$ \\
& $3$  & $54.66$ & $-6.07$  & $64.10$ & $18.94$ \\
& $5$  & $51.57$ & $-9.28$  & $53.50$ & $18.94$ \\
& $10$ & $45.56$ & $-15.62$ & $51.94$ & $18.94$ \\
\hline
\multirow{4}{*}{Rank}
& $1$  & $50.96$ & $-10.49$ & $57.87$ & $13.37$ \\
& $3$  & $54.66$ & $-6.07$  & $64.10$ & $18.94$ \\
& $5$  & $51.54$ & $-9.53$  & $65.17$ & $24.50$ \\
& $10$ & $50.39$ & $-10.26$ & $79.12$ & $37.05$ \\
\hline
\multirow{4}{*}{Buffer Size}
& $100$  & $50.59$ & $-10.13$ & $59.13$ & $18.94$ \\
& $400$  & $54.66$ & $-6.07$  & $64.10$ & $18.94$ \\
& $700$  & $55.39$ & $-4.52$  & $65.05$ & $18.94$ \\
& $1000$ & $56.19$ & $-3.86$  & $81.04$ & $18.94$ \\
\bottomrule
\end{tabular}
\end{table}

}

{
\textbf{Effect of Server-side Recompression.}
To evaluate the effectiveness of DSHA, we consider a variant that retains client-side low-rank compression but removes server-side recompression, directly broadcasting all uploaded curvature factors to participating clients. As shown in Table~\ref{tab:server-side recompression}, DSHA reduces communication costs by $81.6\%$ and per-round runtime by $66.8\%$, while also reducing peak GPU memory from $20.23$ GB to $18.52$ GB. This efficiency improvement comes with a $1.77$ pp decrease in ACC but a $1.57$ pp improvement in BWT, demonstrating that DSHA substantially reduces computational and communication overhead while maintaining competitive learning performance.
\begin{table*}[h]
    \centering
    \caption{Ablation study of curvature correction and server-side recompression on CIFAR-100, comparing their effects on learning performance, computational overhead, peak GPU memory, and communication cost.}
    \label{tab:server-side recompression}
    \resizebox{\textwidth}{!}{%
    \begin{tabular}{lccccc}
        \toprule
         & ACC (\%) $\uparrow$
         & BWT (pp) $\uparrow$
         & Time/round (s) $\downarrow$
         & Peak Memory Usage (GB) $\downarrow$
         & Communication Cost (MB) $\downarrow$ \\
        \midrule
        No Curvature  &  $40.19$ & $-13.94$ & $36.46$ & $12.03$ & $175.89$\\
        Remove recompression
         & $56.58$ & $-7.61$ & $192.92$ & $20.23$ & $1889.275$ \\
        FedGRIT
         & $54.81$ & $-6.04$ & $64.10$ & $18.52$ & $347.88$ \\
        \bottomrule
    \end{tabular}%
    }
\end{table*}

}

\subsection{Computation and communication cost}
We report the results of rounds-to-best and time-per-round for all algorithms in Table~\ref{tab:training_communication}. Compared with other A-FCL methods, \Adaptive performs client-specific weight adaptation, allowing each client to determine its own optimal update direction at each local gradient step instead of following the same weights as other clients. This personalized adaptation strategy enables \Adaptive to achieve the fewest rounds-to-best among all FCL and A-FCL methods. The numerical efficiency comparisons are provided in Table~\ref{tab:training_memory}. The results show that methods performing information exchange and processing entirely on the client side incur communication costs comparable to FedAvg. In contrast, federated multi-objective learning methods require the transmission of multiple task gradients, FedAGC additionally downloads group models, and C-FLAG exchanges $\nabla g(x_t)$ and $\nabla f^\dagger(x_t)$ with the server to determine adaptive learning rates. Similarly, FedGRIT requires additional curvature-information exchange between clients and the server. These additional client-server interactions inevitably introduce communication overhead beyond standard FL.

\begin{table*}[h]
\centering
\caption{Comparison of convergence speed and runtime. We report the \emph{rounds-to-best} and \emph{time-per-round} over all tasks for each dataset. Lower is better for both metrics.}
\label{tab:training_communication}
\resizebox{\textwidth}{!}{%
\begin{tabular}{lcccccccc}
\hline
\multirow{2}{*}{Method}
& \multicolumn{2}{c}{CIFAR-100}
& \multicolumn{2}{c}{ImageNet100}
& \multicolumn{2}{c}{Digit-10}
& \multicolumn{2}{c}{VLCS} \\
\cmidrule(lr){2-3}\cmidrule(lr){4-5}
\cmidrule(lr){6-7}\cmidrule(lr){8-9}

& Rounds(r) & Time(s) 
& Rounds(r) & Time(s) 
& Rounds(r) & Time(s) 
& Rounds(r) & Time(s) \\
\hline

FedAvg
& $226.7$ & $\textbf{27.7}$
& $221.7$ & $\textbf{36.7}$
& $\underline{55.7}$ & $\textbf{68.8}$
& $90.7$ & $\textbf{51.2}$ \\
FedProx
& $220.0$ & $34.7$
& $220.7$ & $39.6$
& $60.0$ & $75.6$
& $83.7$ & $\underline{52.1}$ \\
\hline

FSMGDA
& $245.7$ & $33.4$
& $240.7$ & $42.7$
& $107.7$ & $75.3$
& $88.7$ & $53.7$ \\
FedCMOO
& $237.3$ & $40.3$
& $242.7$ & $47.0$
& $95.0$ & $72.8$
& $78.0$ & $55.6$ \\
\hline

FLwF-2T
& $218.3$ & $39.1$
& $212.0$ & $46.4$
& $84.3$ & $90.9$
& $74.7$ & $52.4$ \\
FedSSI
& $\textbf{162.0}$ & $46.7$
& $216.0$ & $56.7$
& $62.3$ & $141.2$
& $68.7$ & $56.6$ \\
SPECIAL
& $228.3$ & $\underline{28.0}$
& $\underline{199.7}$ & $36.3$
& $64.0$ & $\underline{69.8}$
& $\textbf{57.0}$ & $52.7$ \\
\hline

FedPOCL
& $238.7$ & $86.0$
& $229.3$ & $93.4$
& $98.7$ & $143.1$
& $75.0$ & $65.2$ \\
FedAGC
& $244.0$ & $101.4$
& $229.7$ & $102.1$
& $107.0$ & $150.1$
& $94.0$ & $56.9$ \\
FedBECAME
& $242.3$ & $53.0$
& $221.3$ & $78.4$
& $106.0$ & $102.6$
& $60.7$ & $77.7$ \\

C-FLAG
& $227.0$ & $59.3$
& $224.0$ & $48.6$
& $68.7$ & $92.6$
& $75.3$ & $59.3$ \\
\Adaptive (ours)
& $\underline{206.3}$ & $64.1$
& $\textbf{195.7}$ & $72.9$
& $\textbf{41.0}$ & $131.3$
& $\underline{66.7}$ & $69.2$ \\
\hline
\end{tabular}}
\end{table*}

\begin{table}[h]
\centering
\caption{Numerical efficiency comparison on CIFAR-100. Communication cost is measured per selected client per round, including both upload and download. Peak memory usage refers to peak allocated GPU memory.}
\label{tab:training_memory}
{%
\begin{tabular}{lcc}
\hline
Method
& Communication cost (MB) $\downarrow$
& Peak memory usage (GB) $\downarrow$ \\
\hline
FedAvg
& $85.73$
& $4.40$ \\
FedProx
& $85.73$
& $4.35$ \\
\hline
FSMGDA
& $171.47$
& $6.31$ \\
FedCMOO
& $214.23$
& $6.22$ \\
\hline
FLwF-2T
& $85.73$
& $5.39$ \\
FedSSI
& $85.73$
& $22.06$ \\
SPECIAL
& $85.73$
& $4.40$ \\
\hline
FedPOCL
& $85.73$
& $4.74$ \\
FedAGC
& $128.60$
& $8.51$ \\
FedBECAME
& $85.73$
& $9.18$ \\
C-FLAG
& $257.19$
& $12.65$ \\
{\Adaptive (ours)}
& ${347.88}$
& ${18.52}$ \\
\hline
\end{tabular}}
\end{table}

{
\subsection{Discussion about Hessian-vector Product (HVP) budget}
Algorithm~\ref{alg:Hessian-Lanczos Decomposition}  requires one Hessian-vector product (HVP) per Lanczos iteration, with a maximum of ten HVPs per historical task per participating client at each curvature refresh. The retained rank determines the curvature representation size and reconstruction cost, while the Lanczos iteration budget determines the number of HVP evaluations. Between refreshes, cached curvature factors are reused without additional HVPs. For a model with $d$ parameters and $K-1$ historical tasks, each participating client exchanges $(K-1)(dr+r+d)$ scalars per communication direction at a complete refresh, excluding model transmission and updates. Each historical-task gradient correction requires $\mathcal{O}(dr)$ operations per local step. Thus, communication and reconstruction costs scale linearly with $d$ and $K$ for fixed $r$, while DSHA maintains a fixed-size curvature representation per task regardless of the number of participating clients.
}

\end{document}